\documentclass[submit]{bioRxiv}
\usepackage[numbers,sort&compress]{natbib} 

\usepackage{lipsum}
\usepackage{adjustbox} 
\usepackage{hyperref}
\usepackage{ragged2e} 
\usepackage{longtable} 
\usepackage{array}     
\usepackage{setspace}
\usepackage{multirow} 
\renewcommand{\cite}[1]{\textsuperscript{[\citenum{#1}]}}
\renewcommand{\citep}[1]{\textsuperscript{[\citenum{#1}]}}
\renewcommand{\citet}[1]{\textsuperscript{[\citenum{#1}]}}

\begin{document}
\nolinenumbers
 \leadauthor{Li}

\newcommand{\mytitle}{Learning neural representations for X-ray ptychography reconstruction with unknown probes}

\title{\singlespacing \huge \mytitle}
\shorttitle{\textbf{PtyINR}: \textbf{Pty}chographic \textbf{I}mplicit \textbf{N}eural \textbf{R}epresentation}

\author[1]{Tingyou Li}
\author[2]{Zixin Xu}
\author[3]{Zirui Gao}
\author[3]{Hanfei Yan}
\author[3, \Letter]{Xiaojing Huang}
\author[1, \Letter]{Jizhou Li}
\affil[1]{Department of Electronic Engineering, The Chinese University of Hong Kong, Shatin, N.T., Hong Kong}
\affil[2]{Department of Computer Science and Engineering, The Hong Kong University of Science and Technology, Clear Water Bay, Kowloon, Hong Kong}
\affil[3]{National Synchrotron Light Source II, Brookhaven National Laboratory, Upton, NY, USA}

\date{}

\doublespacing
\maketitle

\begin{corrauthor}
jzli@cuhk.edu.hk (J. Li) and xjhuang@bnl.gov (X. Huang)
\end{corrauthor}

\begin{abstract}
\section*{Abstract}

X-ray ptychography provides exceptional nanoscale resolution and is widely applied in materials science, biology, and nanotechnology. However, its full potential is constrained by the critical challenge of accurately reconstructing images when the illuminating probe is unknown. Conventional iterative methods and deep learning approaches are often suboptimal, particularly under the low-signal conditions inherent to low-dose and high-speed experiments. These limitations compromise reconstruction fidelity and restrict the broader adoption of the technique. In this work, we introduce the Ptychographic Implicit Neural Representation (PtyINR), a self-supervised framework that simultaneously addresses the object and probe recovery problem. By parameterizing both as continuous neural representations, PtyINR performs end-to-end reconstruction directly from raw diffraction patterns without requiring any pre-characterization of the probe. Extensive evaluations demonstrate that PtyINR achieves superior reconstruction quality on both simulated and experimental data, with remarkable robustness under challenging low-signal conditions. Furthermore, PtyINR offers a generalizable, physics-informed framework for addressing probe-dependent inverse problems, making it applicable to a wide range of computational microscopy problems.

\end{abstract}


\section*{Main}\label{s:introduction}
X-ray ptychography is a high-resolution, lensless imaging technique that reconstructs the complex-valued transmission function of an object with extended dimensions from coherent diffraction patterns. The reconstructed amplitude and phase functions of the object represent the imaginary and real parts of the material’s refractive index. This technique has been successfully applied across various fields, including materials science~\cite{grote_imaging_2022,diaz_characterization_2014}, biology~\cite{hemonnot_imaging_2017,zhu_measuring_2016,gorecki_ptychographic_2023,rose_quantitative_2018,piazza_revealing_2014}, and nanotechnology~\cite{holler_high-resolution_2017}, due to its ability to achieve nanoscale resolution beyond the limitations of conventional optics~\cite{pfeiffer_x-ray_2018,miao2025computational}. 

\begin{figure}[!t]
    \centering
    \includegraphics[width=\textwidth, clip]{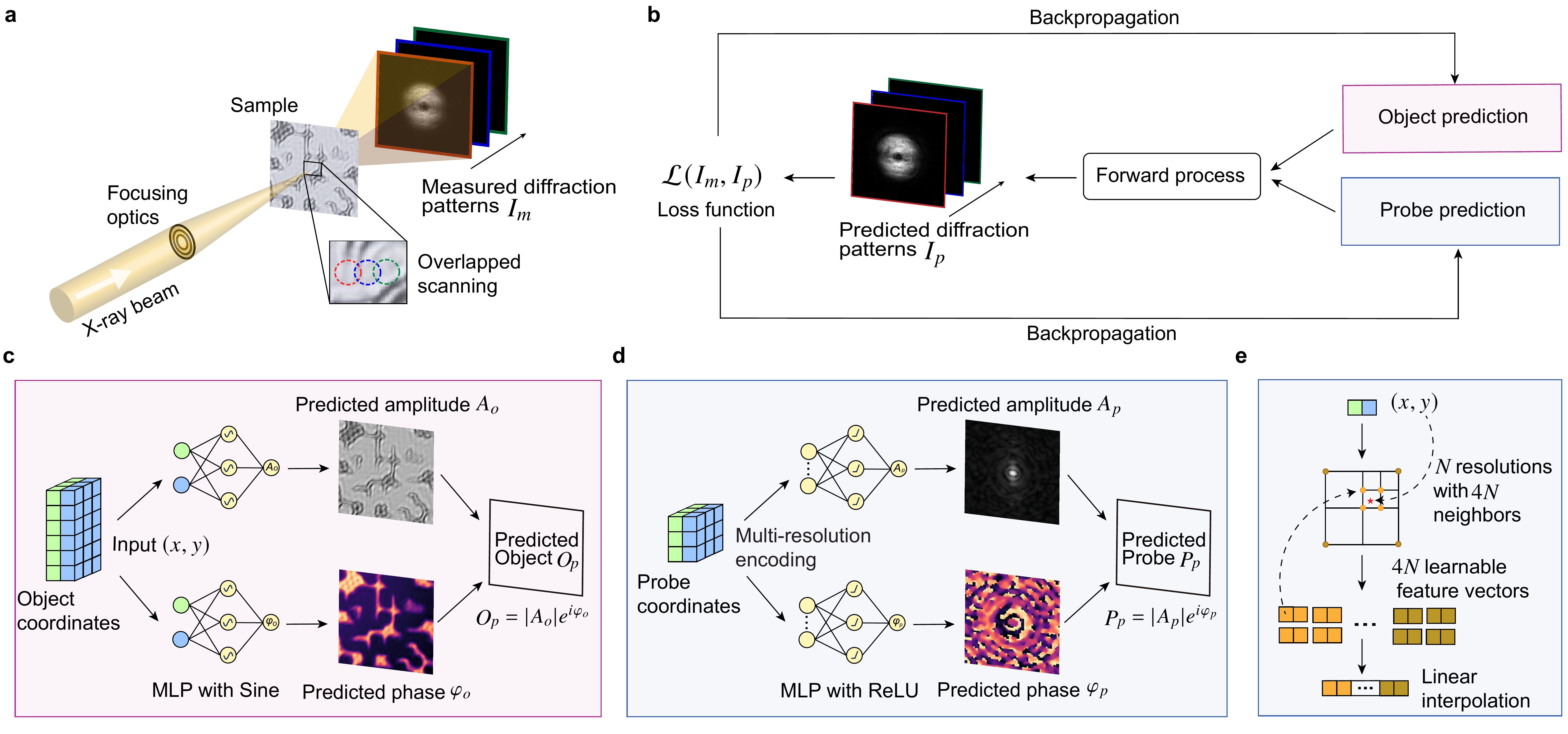}  
    \caption{\textbf{Overview of the PtyINR framework for joint object-probe reconstruction for X-ray ptychography.}
        (\textbf{a}) Schematic of the X-ray ptychography setup, where coherent X-rays are focused onto a sample using a zone plate~\cite{pattammattel2020high} or multilayer Laue lens~\cite{yan_multimodal_2018}. The sample is scanned overlappingly, and diffraction patterns $I_m$ are measured at each position.
        (\textbf{b}) The reconstruction pipeline that jointly predicts the object and probe by minimizing the difference between measured and predicted diffraction patterns ($I_m$ and $I_p$) via backpropagation.
        (\textbf{c}) Neural network architecture for object prediction. Spatial coordinates of the object are input to two separate MLPs (multi-layer perceptrons) with sine activation~\cite{sitzmann_implicit_2020} to predict the amplitude $A_o$ and phase $\varphi_o$, which are subsequently combined into the predicted object $O_p$.
        (\textbf{d}) Neural network architecture for the probe prediction, using multi-resolution encoding in (\textbf{e}) and MLPs with ReLU activation to predict the amplitude $A_p$ and phase $\varphi_p$ of the probe $P_p$.}
    \label{illustrate}
\end{figure}

Despite its broad applicability, reconstructing images from ptychographic data remains a significant challenge. In X-ray ptychography, a coherent and focused X-ray beam is scanned across the sample at overlapping positions (Fig.~\ref{illustrate}a). At each scan point, the X-ray wave interacts with the sample, generating an exit wave that encodes both amplitude and phase information. However, in the experiments, only the intensity of the exit wave propagating to the detector is measured, and the phase information is lost. This results in a non-linear inverse problem known as phase retrieval, which adds substantial computational complexity. Additionally, the exit wave encodes information about both the object and the probe (illumination function). The simultaneous recovery of these two components introduces an intrinsic duality, where errors in one reconstruction propagate and affect the other, leading to interdependent inaccuracies. Unlike wavefront aberrations, which can be efficiently modeled using compact representations such as Zernike polynomials, the probe in ptychography typically lacks a low-dimensional parametric form. In many cases, the probe’s values are treated as independent and measured separately, further increasing the dimensionality of the problem and compounding the computational burden.

The conventional iterative algorithms have tried to jointly reconstruct both the object and probe. These include stochastic methods like the extended Ptychographic Iterative Engine (ePIE)~\cite{maiden_improved_2009,maiden_further_2017}, projection-based approaches such as the Difference Map (DM)~\cite{thibault_probe_2009} and Relaxed Averaged Alternating Reflections (RAAR)~\cite{luke_relaxed_2004}, and recent techniques like the Weighted Average of Sequential Projections (WASP)~\cite{maiden_wasp_2024} and Accelerated Proximal Gradient (APG)~\cite{yan_ptychographic_2020} and so on~\cite{zhai_projected_2023,odstrcil_iterative_2018}. Notably, a latest study~\cite{bui_stochastic_2024} further enhanced the ADMM algorithm to better accommodate large-scale ptychography. To address computational bottlenecks, GPU-accelerated libraries~\cite{enders_computational_2016,mandula_pynxptycho_2016,marchesini_sharp_2016,nashed_parallel_2014} have significantly reduced reconstruction times, enabling practical and accessible high-resolution imaging. While these iterative methods, such as ePIE, DM, and RAAR, have demonstrated considerable success in X-ray ptychographic reconstruction, they often suffer from limitations, including the need for sufficient overlap rates between scan positions, sensitivity to experimental noise, and a strong dependence on careful initialization of both the object and the probe~\cite{cherukara_ai-enabled_2020}.

Deep learning has emerged as a promising alternative, with supervised models like PtychoNN~\cite{cherukara_ai-enabled_2020,babu_deep_2023}, PtychoNet~\cite{guan_ptychonet_2019}, and the transformer-based PtychoDV~\cite{gan_ptychodv_2024} achieving rapid reconstructions through learned mappings between diffraction patterns and objects. While these U-Net-based models demonstrate enhanced noise resistance and robustness under low-overlap acquisition conditions compared to iterative methods, they suffer from a critical limitation rooted in their dependence on extensive paired training datasets, which are challenging to acquire experimentally. These datasets are typically generated through conventional algorithm-based pre-reconstruction (e.g., ePIE or DM), thus failing to capture the full complexity of real-world probe variations, detector noise, probe instability, and sample heterogeneity. More critically, these supervised methods also exhibit practical risk to overfit to training-specific artifacts (e.g., ePIE's edge ringing) rather than learning diffraction physics, potentially generating misleading reconstructions when encountering unexpected experimental scenarios.  In recent years, self-supervised approaches have gained attention for ptychographic reconstruction. One representative of such methods leverages auto-differentiation (AD) back-ended by PyTorch~\cite{paszke_pytorch_2019} to jointly optimize both the object and the probe by minimizing the discrepancy between predicted and measured diffraction patterns ~\cite{kandel_using_2019,du_three_2020,du_adorym_2021,wu_dose-efficient_2024}. These methods parameterize the object and probe as differentiable matrices and recover them through gradient-based optimization. While AD frameworks offer flexibility and compatibility with modern optimization tools, their reliance on discrete pixel-wise representations introduces critical bottlenecks. The grid-based parameterization struggles to adapt to irregular scan geometries or partial overlaps, generates aliasing artifacts that degrade resolution, and can be sensitive to noise, where perturbations in the measurements can lead to instability or amplification of reconstruction errors due to the rigid and discrete nature of the underlying representation. Consequently, AD-based approaches are ill-suited for real-world applications such as low-dose imaging of radiation-sensitive specimens, dynamic \textit{in situ} processes, or experiments employing sparse or adaptive sampling protocols.

More recently, neural network-based self-supervised methods, such as PINN~\cite{hoidn_physics_2023} and PtyNet~\cite{pan_efficient_2023}, have been proposed to address this problem. These approaches use architectures like U-Net to learn the mapping from diffraction patterns to object reconstructions, defining the loss based on measurement consistency rather than requiring ground-truth labels. Although they provide a continuous and more expressive representation of the object, a key limitation is the assumption of a known probe, which is often difficult to obtain accurately in advance. Therefore, in practice, these methods typically depend on iterative algorithms to estimate the probe, meaning their overall performance remains largely constrained by the limitations of such iterative procedures.


To address the above limitations, we proposed a novel reconstruction framework named PtyINR, which leverages Implicit Neural Representations (INRs)~\cite{sitzmann_implicit_2020} and ptychographic physics to jointly solve the challenging reconstruction problem with unknown probes. More specifically, PtyINR represents both the object and the probe as continuous functions that are parameterized by multilayer perceptrons (MLPs), with spatial coordinates serving as the input to each network. In PtyINR, we applied asymmetric model architectures tailored to the distinct characteristics of the object and the probe, and introduced a novel loss function to enhance the stability of PtyINR training across diverse scenarios. Prior to our work, INRs had demonstrated success across multiple computational imaging domains—notably Fourier ptychography~\cite{zhou_fourier_2023}, adaptive optics in wide-field microscopy~\cite{kang_coordinate-based_2024}, and tomographic reconstruction~\cite{liu_recovery_2022,sun_coil_2021}.

PtyINR enables the simultaneous recovery of both the object and the probe using a unified self-supervised neural network. Besides, compared to existing methods, it achieves state-of-the-art reconstruction performance. Simulation and experimental results demonstrate that PtyINR enables accurate ptychographic reconstruction under low-dose conditions, such as short exposure times or low scanning overlap ratios, making it particularly suitable for imaging radiation-sensitive materials. Moreover, our framework exhibits strong robustness to the measurement noise, ensuring reliable performance under realistic experimental conditions. Therefore, the development of PtyINR eliminates the dependence on prior constraints, well-informed probe initialization, offering a more efficient and accurate solution to X-ray ptychography reconstruction. Furthermore, PtyINR can provide a generalizable framework that can be readily adapted for other imaging modalities requiring accurate probe recovery, such as near-field ptychography~\cite{stockmar_near-field_2013} and Bragg ptychography~\cite{li2021revealing}. Together, these advancements position PtyINR as a powerful and versatile tool for next-generation coherent X-ray imaging, with broad implications for high-resolution, low-dose, and noise-resilient nanoscale material characterization with computational microscopy.


\section*{Results}\label{s:results}

As shown in Fig.~\ref{illustrate}b-e, the proposed PtyINR jointly reconstructs both the object and the illumination probe directly from the measured diffraction patterns without prior training (see the network architectural details in the~\hyperref[s:methods]{Methods} section). We conduct comprehensive experiments on both simulation across various scenarios (e.g. noise degradation, scanning overlap ratios, and exposure time) and real data acquired from the Hard X-ray Nanoprobe (HXN) beamline of the National Synchrotron Light Source II (NSLS-II) at Brookhaven National Laboratory. Our method consistently outperform existing techniques, especially when operating with sparse or low signal-to-noise data, demonstrating superior robustness and reconstruction quality.




\vspace{-3pt}
\subsection*{Simulations on Different Scanning Overlap Ratios}

\begin{figure}[!t]
    \centering
    \includegraphics[width=\textwidth, height=\textheight, keepaspectratio]{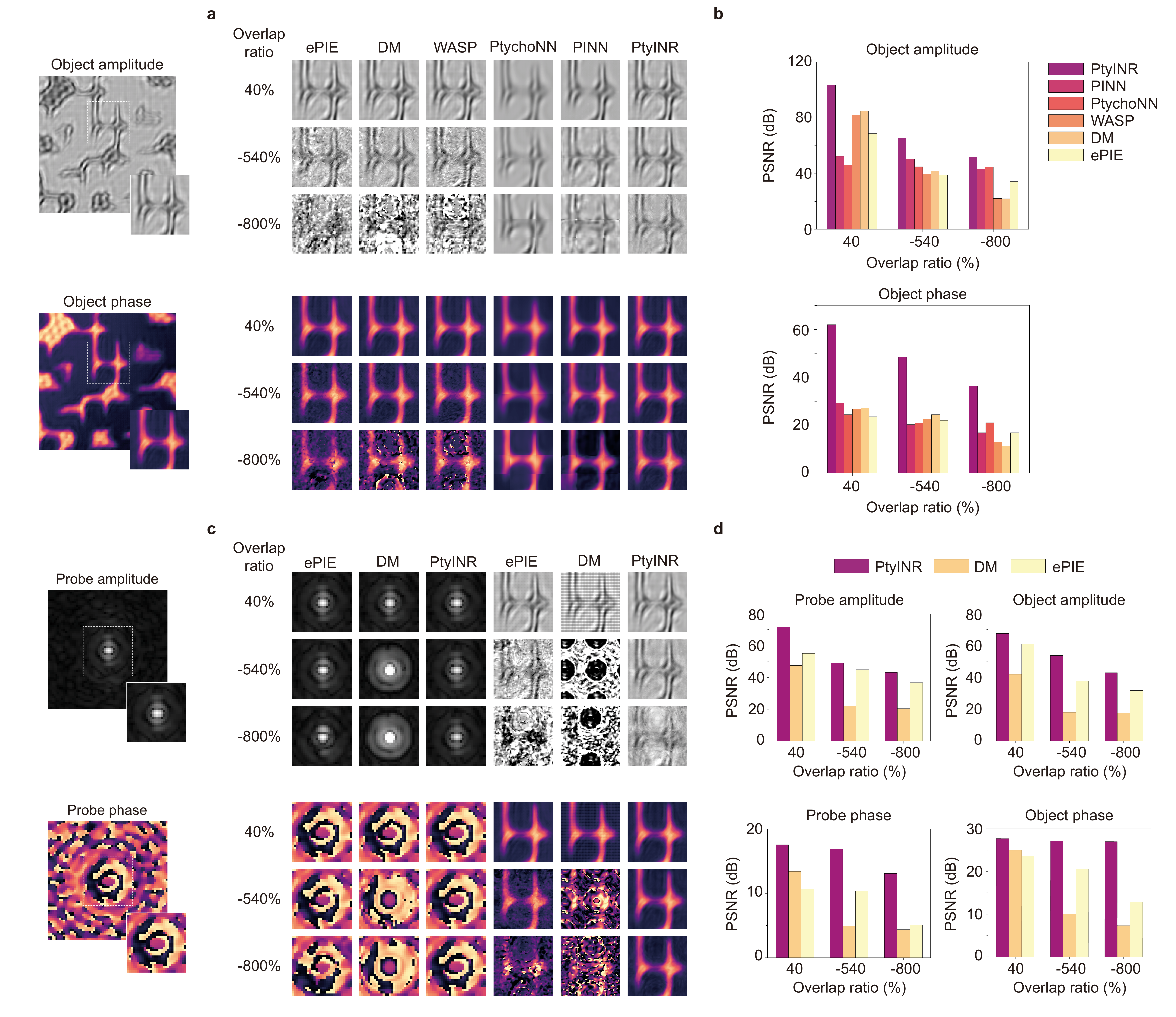}  
    \caption{\textbf{Object and probe reconstruction accuracy across scanning overlap ratios.}
        Amplitude (grayscale) and phase (warm colormap) representations under different experimental configurations (ground truth shown left).
        (\textbf{a}) Probe-known condition: Object reconstructions at 40\%, -540\%, and -800\% overlap ratios using ePIE, DM, WASP, PtychoNN, PINN, and PtyINR.
        (\textbf{b}) Quantitative comparison of object amplitude/phase reconstruction quality via peak signal-to-noise ratio (PSNR) across methods.
        (\textbf{c}) Probe-unknown condition: Reconstructed probe profiles (left) and corresponding objects (right) from ePIE, DM, and PtyINR.
        (\textbf{d}) PSNR evaluation for reconstructed probe and object amplitude/phase components.}
    \label{simulate_overlap}
    \vspace{6pt}
\end{figure}
To evaluate the reconstruction performance under different overlap conditions, we compare the proposed PtyINR with five representative ptychographic algorithms: ePIE~\cite{maiden_improved_2009}, DM~\cite{thibault_probe_2009}, WASP~\cite{maiden_wasp_2024}, PtychoNN~\cite{cherukara_ai-enabled_2020}, PINN~\cite{hoidn_physics_2023}. Reconstructions are assessed both qualitatively and quantitatively under conditions when the probe is known and unknown across three probe overlap ratios $(1-\frac{\text{step size}}{\text{probe diameter}})\times100\%$: 40\%, -540\%, and -800\%. The diffraction patterns are simulated using raster step scanning, without the inclusion of noise or positional distortions. The ground truth probe is extracted from a previously reconstructed probe obtained from experimental datasets. And the probe size is estimated based on its intensity profile using the full width at half maximum (FWHM), as shown in Fig.~\ref{fig:supplementary_fig0}a. It is noteworthy that although the overlap ratio is negative, due to the probe being significantly smaller than the probe array, the scanning process still includes overlapping regions, as illustrated in Fig.~\ref{fig:supplementary_fig0}b. Specifically, in our experimental configuration, nominal overlaps of 40\%, -540\%, and -800\% correspond to approximately 95\%, 50\%, and 30\% actual overlap of the full probe array, respectively. Consequently, successful object recovery requires reconstructing not only the central probe but also the remainder of the probe array. In this study, the iterative reconstruction methods are employed using Ptypy~\cite{enders_computational_2016}, a highly optimized and widely used library for ptychographic imaging.  To ensure optimal initialization, the iterative methods are provided with a probe derived from prior experimental reconstructions conducted under similar conditions.  

Fig.~\ref{simulate_overlap}a shows the reconstructed object amplitude (gray images) and object phase (colored images) of different methods under different overlap ratios when the probe is known. The left-most column presents the ground truth. As the overlap decreases, traditional optimization-based methods (ePIE, DM, WASP) and even learning-based method PINN exhibit notable degradations in both amplitude and phase, particularly in the -800\% overlap regime. Although PtychoNN demonstrate consistent and reliable performance across varying overlap ratios, it fail to preserve fine details of the original object, particularly evident in the edge regions. In contrast, PtyINR yields consistently high-quality reconstructions across all overlap conditions, demonstrating improved robustness and generalization.

Quantitative comparisons in Fig.~\ref{simulate_overlap}b report the peak signal-to-noise ratio (PSNR) for the object reconstructions when the probe is known. PtyINR achieves significantly higher PSNR in both amplitude and phase, particularly under reduced overlap, highlighting its superior fidelity in reconstructing fine structural details. In addition to these methods, we also compare against the auto-differentiation (AD)~\cite{du_adorym_2021} and relaxed averaged alternating reflections (RAAR) in Fig.~\ref{fig:supplementary_fig1}. It is notable that AD parameterizes the object as a gradient-required matrix rather than using the continuous neural representation employed by PtyINR. This comparison also serves as an ablation study to validate the effectiveness of our neural network-based object representation. As shown in Fig.~\ref{fig:supplementary_fig1} through both qualitative and quantitative evaluations, the AD method initially exhibits performance comparable to PtyINR when the probe was known. However, its performance declines rapidly as the overlap ratio decreases. When the overlap ratio drops to -800\%, AD's performance becomes nearly identical to that of RAAR~\cite{luke_relaxed_2004} and is significantly worse than that of PtyINR. 

Moreover, we evaluates the recovery performance of PtyINR against the other two classical iterative methods when the probe is unknown. As shown in Fig.~\ref{simulate_overlap}c, which presents both the recovered probe and object, PtyINR consistently outperforms ePIE and DM, maintaining high-quality reconstructions even at a -800\% overlap ratio. The corresponding PSNR analysis in Fig.~\ref{simulate_overlap}d further supports these observations; PtyINR systematically achieves the highest PSNR across all reconstruction categories, including object amplitude, object phase, probe amplitude, and probe phase.

Finally, we compares PtyINR against algorithms of RAAR, WASP, and AD in Fig.~\ref{fig:supplementary_fig2} and Fig.~\ref{fig:supplementary_fig3} when the probe is unknown. This comparison highlights the advantages of PtyINR's neural representation and optimization strategy across different reconstruction paradigms. In Fig.~\ref{fig:supplementary_fig2}, RAAR, WASP, and AD fails to recover the object amplitude when the overlap ratio decreases to -540\%. At an overlap ratio of -800\%, these methods fails entirely, whereas PtyINR successfully reconstructs the object phase with high fidelity and preserves an approximate structure of the object amplitude. Furthermore, PtyINR consistently achieves more accurate probe reconstructions across different overlap ratios as shown in Fig.~\ref{fig:supplementary_fig3}. These findings are further supported by the PSNR histograms.

PtyINR generally outperforms other deep learning-based methods when the probe is known, which can be attributed to the strong expressive power of our model architectures. Moreover, PtyINR demonstrates superior performance compared to both iterative methods and AD-based approaches under both known and unknown probe conditions. This advantage may stem from the fact that conventional methods typically represent the object as a discrete set of pixels, whereas PtyINR models it as a continuous function. As a result, PtyINR exhibits greater robustness to sharp artifacts introduced by insufficient or degraded data. However, this does not imply that PtyINR is incapable of learning sharp, high-frequency details when they genuinely exist in the ground truth. Accurate recovery of such details requires a sufficient amount of input information to allow the model to adapt its representation accordingly. In the case of limited data, PtyINR tends to learn a smoother approximation, which can be considered optimal given the inherent trade-off between resolution and data availability.

\begin{figure}[!t]
    \centering    \includegraphics[width=\textwidth, height=\textheight, keepaspectratio]{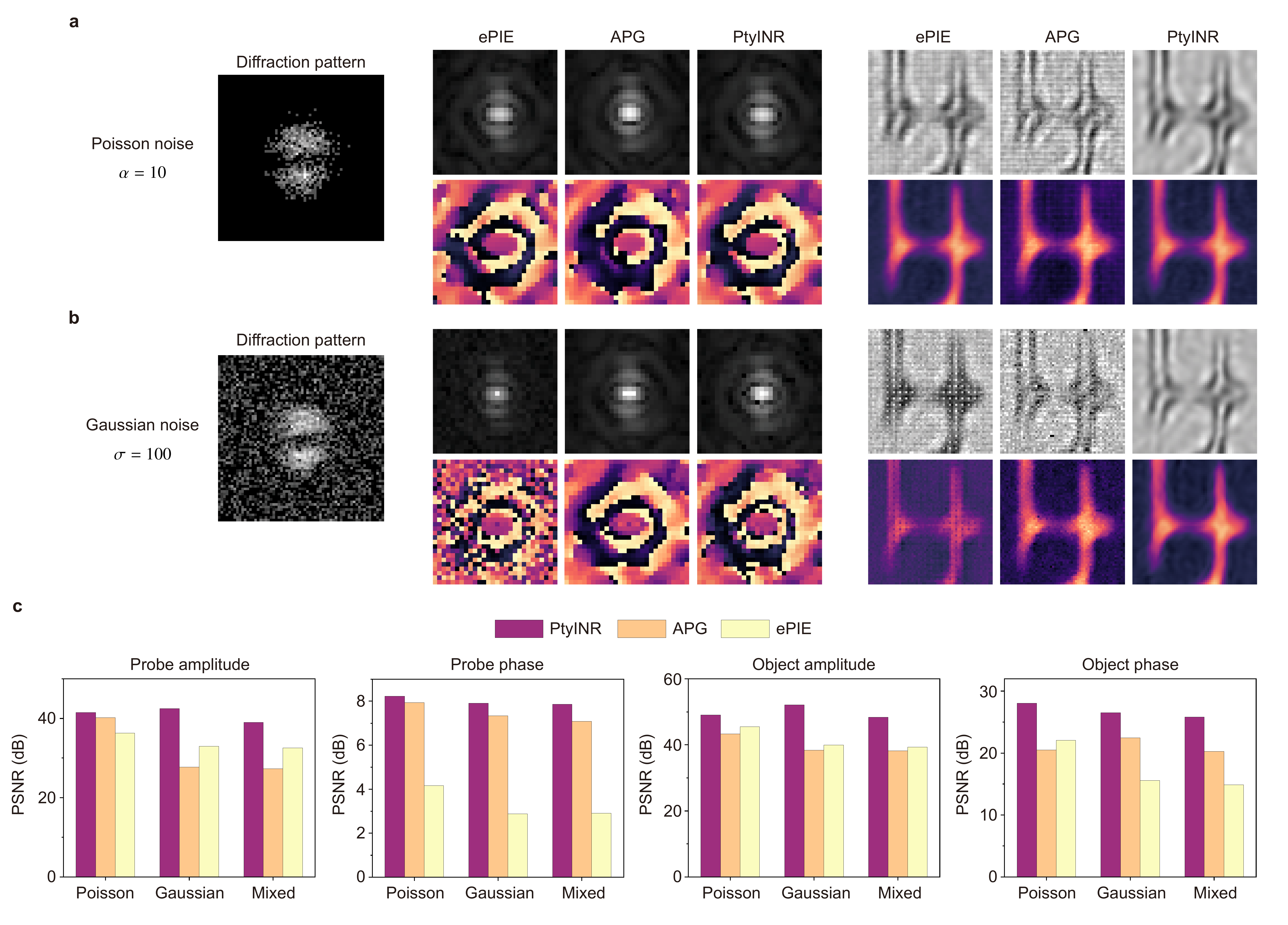}
    \vspace{-10pt}
    \caption{\textbf{Method robustness against noise in probe and object reconstructions.}
        Amplitude (grayscale) and phase (warm colormap) reconstructions under (a) Poisson noise and (b) Gaussian noise, showing probe (left) and object (right) results for ePIE, APG, and PtyINR. (\textbf{c}) Quantitative peak signal-to-noise ratio (PSNR) comparisons under Poisson, Gaussian, and mixed noise for (left to right): probe amplitude, probe phase, object amplitude, and object phase. PtyINR demonstrates superior noise resilience, achieving the highest PSNR values in all categories.}
    \label{simulate_noise}
\end{figure}
In contrast, iterative and AD-based methods are inherently designed to recover pixel-level detail regardless of data sufficiency. Consequently, when the data is abundant, such as in the case of a 40\% overlap scanning ratio, there is no significant performance difference between PtyINR and these methods. However, as the overlap ratio decreases, the radiation-dose efficiency of PtyINR becomes increasingly evident, outperforming other methods by achieving better reconstructions with fewer exposures.

\vspace{-10pt}
\subsection*{Simulations on Different Noise Levels}

Nonetheless, in practical scenarios, acquiring entirely noise-free data is rarely feasible. Therefore, to assess the robustness of PtyINR against measurement noise, we benchmarks its performance against ePIE~\cite{maiden_improved_2009} and APG~\cite{yan_ptychographic_2020} under three noise conditions: Poisson noise, Gaussian noise, and a mixed noise model. It is noticeable that APG is specifically designed to handle noisy data via a maximum-likelihood formulation. Representative reconstructions of both the probe and object are shown in Fig.~\ref{simulate_noise}. Under Poisson noise (Fig.~\ref{simulate_noise}a), PtyINR produces visibly cleaner and more accurate reconstructions in both amplitude and phase compared to conventional algorithms. This trend persists under Gaussian noise (Fig.~\ref{simulate_noise}b), where PtyINR continues to outperform ePIE and APG in preserving fine structural details. In particular, ePIE is significantly affected by the Gaussian noise.

\begin{figure}[!t]
    \centering
    \includegraphics[width=0.85\textwidth, height=0.85\textheight, keepaspectratio]{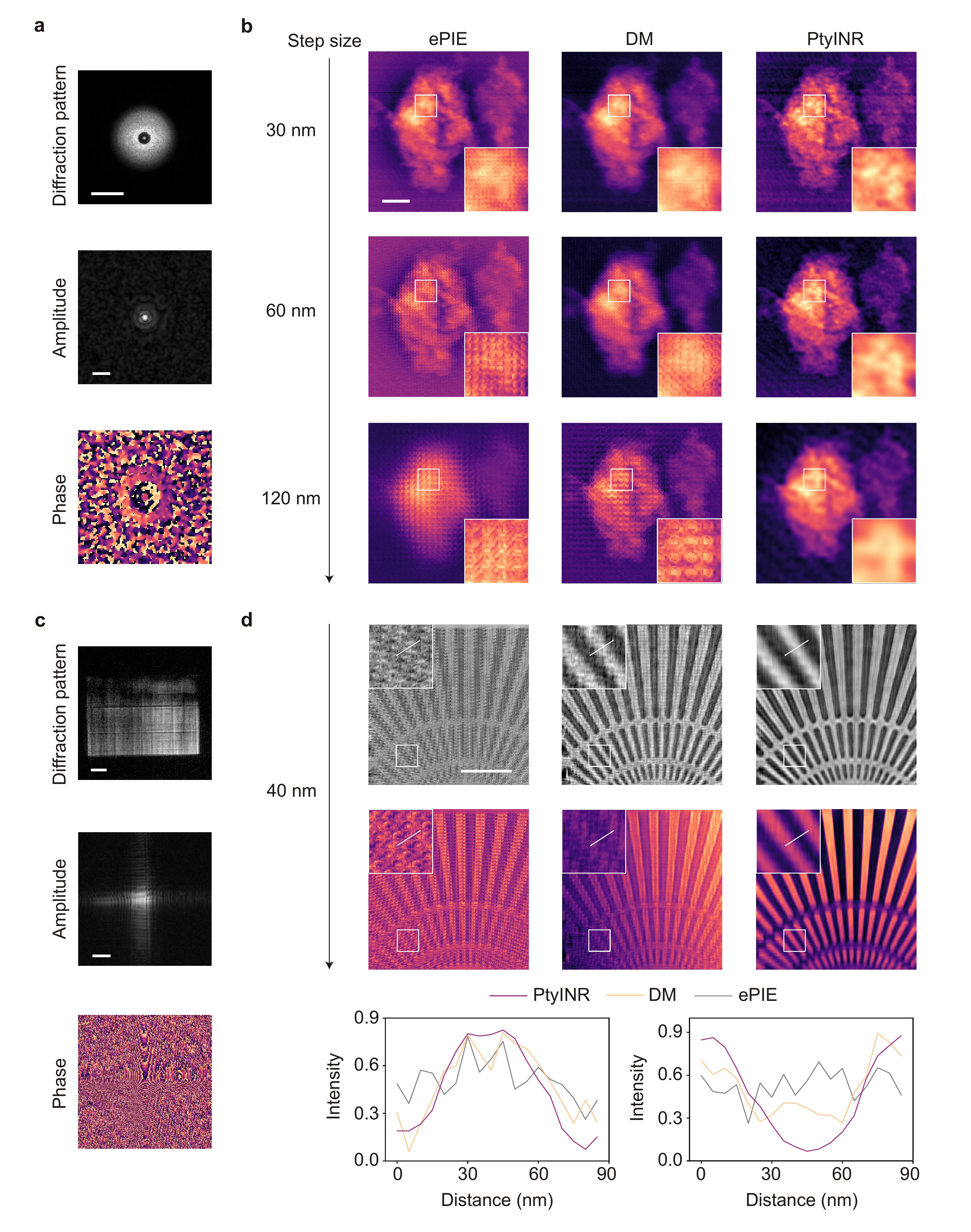}  
    \caption{\textbf{Comparison of reconstruction performance across different focusing systems and methods.}  
    Gray-scale images show amplitude; warm-colored images show phase. Top three rows: FZP system; bottom rows: MLL system. For better visualization effect, the phase color bar is not reversed.
(\textbf{a}, \textbf{c}) Measured diffraction patterns, reference probe amplitude, and phase for FZP and MLL setups. Scale bars: 1.5 mm (diffraction patterns), 150 nm (probes).
(\textbf{b}) Phase reconstructions of a LiCo$\text{O}_2$ sample using FZP at varying scan steps (30 nm, 60 nm, 120 nm) with ePIE, DM, and PtyINR. Insets display magnified regions of interest with a 3× zoom factor. Scale bar: 600 nm.
(\textbf{d}) Amplitude (top) and phase (middle) reconstructions of a Siemens star with MLL at 40 nm defocus, showing improved fidelity with PtyINR. Bottom: line profiles comparing reconstructed amplitude (left) and phase (right), highlighting enhanced contrast and resolution of PtyINR. Scale bar: 600 nm.
}
    \label{real_overlap}
    \vspace{10pt}
\end{figure}
Quantitative comparisons using PSNR in (Fig.~\ref{simulate_noise}c) further support these observations. Across all noise types, PtyINR consistently achieves the highest PSNR values for both probe amplitude and phase, as well as object amplitude and phase, indicating superior fidelity in both probe and object reconstructions. Additional results presented in Fig.~\ref{fig:supplementary_fig4} and Fig.~\ref{fig:supplementary_fig5} also demonstrate that PtyINR outperforms DM, RAAR, WASP, and AD in both object and probe recovery. WASP and AD appear to be particularly susceptible to Gaussian noise, while RAAR and DM exhibit greater robustness. The quantitative analysis in Fig.~\ref{fig:supplementary_fig4} and Fig.~\ref{fig:supplementary_fig5} further validates PtyINR’s superior performance compared to these methods.

\vspace{-10pt}
\subsection*{Experimental Datasets with Different Scanning Step Sizes}

To assess the robustness of PtyINR under varying scanning geometries and step sizes on the experimental dataset, we conduct comparative reconstructions using both a nanostructure LiCo$\text{O}_2$ sample and a Siemens star test pattern (Fig.~\ref{real_overlap}). Fig.~\ref{real_overlap}a-b correspond to the LiCo$\text{O}_2$ dataset, reconstructed under conventional circular probe illumination from a Fresnel zone plate (FZP) with scanning step sizes of 30 nm, 60 nm, and 120 nm, respectively. The bottom panels (Fig.~\ref{real_overlap}c-d) present results on the Siemens star pattern acquired using a multilayer Laue lens (MLL) focusing system with a fixed step size of 40 nm. Fig.~\ref{real_overlap} last row shows intensity line profiles extracted from the Siemens pattern, enabling a direct comparison of edge fidelity. Both datasets were acquired using fly-scan mode \cite{huang_fly-scan_2015}.


Across all step sizes, PtyINR demonstrates superior reconstruction quality relative to ePIE and DM. In the LiCo$\text{O}_2$ sample reconstructions (Fig.~\ref{real_overlap}a-b, top to bottom), PtyINR significantly suppresses the periodic artifacts that become increasingly prominent in ePIE and DM as the step size increases. This is particularly evident at the 120 nm step size, where PtyINR maintains coherent structural features, whereas the other methods exhibit pronounced grid-like artifacts and degraded contrast. Additional comparisons with RAAR, WASP, and AD are provided in Fig.~\ref{fig:supplementary_fig6}. These methods show a similar trend: as the step size increases, artifacts become more dominant, further distinguishing PtyINR’s robustness. Insets in each panel highlight regions of interest, showcasing PtyINR’s superior preservation of fine recovered features. The probes recovered by these methods are shown in Fig.~\ref{fig:supplementary_fig7}. The recovered amplitude images of the LiCo$\text{O}_2$ sample are excluded from the analysis, as they are found to be non-informative.

For the Siemens star pattern reconstructions (Fig.~\ref{real_overlap}c-d and Fig.~\ref{fig:supplementary_fig9}) with a 40 nm scanning step size, PtyINR consistently outperforms the baseline methods. The reconstructions demonstrate that PtyINR is capable of resolving finer radial lines and maintaining smooth transitions at the edges of high-contrast features. The amplitude and phase images exhibit fewer ringing artifacts and better-defined edges, indicating improved spatial resolution. The corresponding line profiles (Fig.~\ref{real_overlap}d and Fig.~\ref{fig:supplementary_fig9}) further quantify these improvements: PtyINR yields smoother and more physically consistent intensity distributions, with reduced oscillations and sharper feature delineation compared to the other algorithms.

Additional results using 10 nm and 30 nm step sizes with Siemens star pattern are presented in Figs.~\ref{fig:supplementary_fig8},~\ref{fig:supplementary_fig9} and~\ref{fig:supplementary_fig10}, further validating the robustness of PtyINR across a range of scanning conditions and overlap ratios. Notably, PtyINR consistently delivers high-fidelity reconstructions across different probe geometries, suggesting reduced sensitivity to variations in spatial sampling density and scanning overlap. This adaptability is critical for practical applications, where experimental constraints often limit the achievable overlap ratio or impose non-ideal scan patterns.

While PtyINR continues to outperform alternative methods, the performance gap is smaller on experimental data compared to simulated data. For example, in simulated settings, PtyINR can recover high-fidelity phase information using only 1\% of the original data, whereas in experimental data, it reconstructs only an approximate object shape under similarly low overlap conditions, as shown in Figure~\ref{real_overlap}b. This discrepancy is likely due to the loss of high-frequency information during acquisition, since experimental diffraction patterns are typically cropped to retain only the central region, excluding noisy or low-intensity edges. Additionally, experimental data are affected by various noise sources and inaccuracies in recorded positions and system parameters, making the reconstruction task inherently more challenging.

\begin{figure}[!t]
    \centering
    \includegraphics[width=\textwidth, height=\textheight, keepaspectratio]{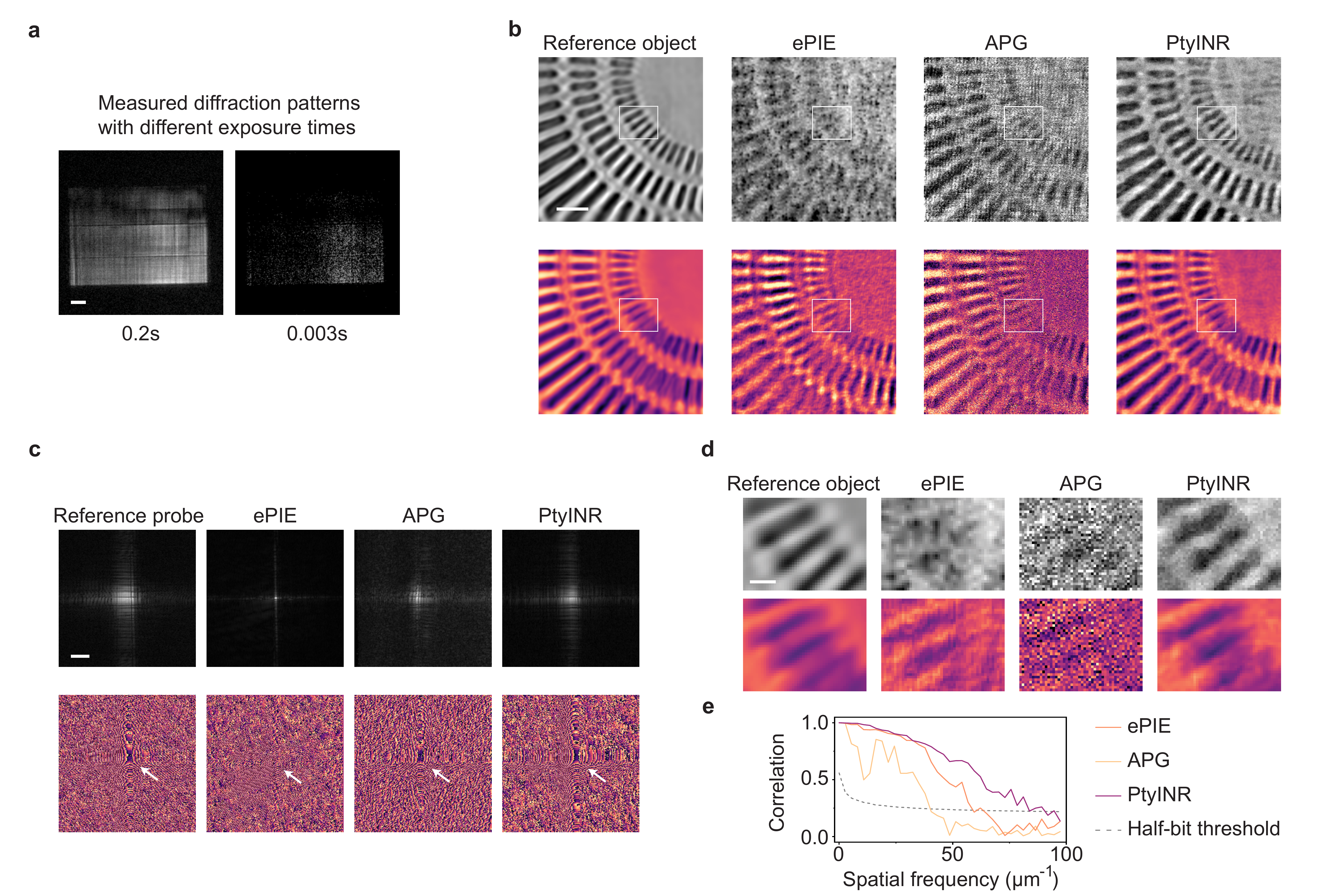}  
    \vspace{-20pt}
    \caption{\textbf{Exposure time-dependent reconstruction fidelity.}
        Amplitude (grayscale) and phase (warm colormap) results compare 0.2s (high-dose reference) and 0.003s (low-dose) conditions.
\textbf{(a)} Diffraction patterns at both exposure times. Scale bar for the diffraction patterns: 1.5 mm.
\textbf{(b)} Low-dose object reconstructions (amplitude top, phase bottom) versus high-dose reference, showing PtyINR’s superior noise suppression. Scale bar: 200 nm.
\textbf{(c)} Probe recovery: High-dose reference vs. low-dose reconstructions (amplitude top, phase bottom). Scale bar: 150 nm.
\textbf{(d)} Magnified regions (white boxes in \textbf{b}) highlighting PtyINR’s retention of fine features. Scale bar: 50 nm.
\textbf{(e)} Fourier ring correlation (FRC) analysis of the object phase at the half-bit FRC threshold.}
    \label{real_exposure}
\end{figure}

\vspace{-10pt}
\subsection*{Experimental Datasets under Extremely Low-Exposure Time}

To evaluate the robustness and reconstruction performance of PtyINR under noisy experimental conditions, we conduct comparative benchmarking against ePIE, APG (Fig.~\ref{real_exposure}), and additional methods presented in Figs.~\ref{fig:supplementary_fig11} and~\ref{fig:supplementary_fig12}. Diffraction patterns were acquired at two exposure times (Fig.~\ref{real_exposure}a): a standard 0.2s exposure yielding a high signal-to-noise ratio with a maximum around 60 photons per detector pixel, and a significantly reduced 0.003s exposure to achieve a low-photon-count regime with a maximum around 5 photons per detector pixel. As expected, the shorter exposure resulted in sparse, noise-dominated measurements, highlighting the challenges faced by conventional algorithms in practical scenarios involving radiation-sensitive samples.

Reconstruction results for the object and probe under these two conditions are presented in Fig.~\ref{real_exposure}b-c and Fig.~\ref{fig:supplementary_fig11}. At the 0.2s exposure time, all methods perform comparably well, as demonstrated in Fig.~\ref{fig:supplementary_fig11}. However, under 0.003s exposure conditions, the object amplitude reconstructions (Fig.~\ref{real_exposure}b, top row) reveal that PtyINR produces images with markedly improved contrast and structural clarity compared to ePIE and APG, both of which suffer from pronounced noise artifacts and spatial blurring. In Fig.~\ref{fig:supplementary_fig12}, RAAR and AD even fail to reconstruct the object amplitude effectively under these conditions. The object phase reconstructions (Fig.~\ref{real_exposure}b, bottom row and Fig.~\ref{fig:supplementary_fig12}) further underscore PtyINR’s superiority, demonstrating significantly reduced phase noise and more accurate retrieval of phase details. Zoomed-in regions (Fig.~\ref{real_exposure}d and Fig.~\ref{fig:supplementary_fig12}) reveal that PtyINR preserves fine spatial features and edge sharpness, which are largely degraded or lost in the reconstructions produced by the other methods.

The reconstructed probe fields (Fig.~\ref{real_exposure}c and Fig.~\ref{fig:supplementary_fig12}) further highlight the advantages of PtyINR. While ePIE and APG produce diffuse and suboptimal probe amplitude distributions, PtyINR successfully recovers a probe profile that closely matches the reference probe obtained under long exposure conditions, exhibiting improved shape fidelity and greater intensity localization. The corresponding phase maps, annotated with directional arrows, further demonstrate PtyINR’s ability to reconstruct the probe phase with higher quality and reduced structural noise, confirming the algorithm’s robustness in recovering both object and illumination fields even in the presence of severe measurement noise. It is noteworthy that AD, as shown in Fig.~\ref{fig:supplementary_fig12}, also demonstrates relatively strong performance in probe recovery, particularly in reconstructing the probe phase. However, the recovered probe by AD does not match the fidelity achieved by PtyINR. 

Additionally, to quantitatively assess reconstruction fidelity, we compute the spatial frequency correlation (FRC)~\cite{koho_fourier_2019} of object phase reconstructions (Fig.~\ref{real_exposure}e), comparing PtyINR with the two best-performing baseline algorithms, ePIE and APG. Based on the half-bit correlation threshold criterion, PtyINR achieves superior spatial resolution in the reconstructed object phase compared to other methods, demonstrating its ability to preserve higher correlation values at extended spatial frequencies.

To further evaluate the robustness of PtyINR to Poisson noise under different experimental settings, we conduct additional experiments at reduced exposure times of \SI{0.01}{\second} and \SI{0.001}{\second} using larger scanning step sizes. The results are shown in Fig.\ref{fig:supplementary_fig_add}, where we compare PtyINR with APG. Under the lower-noise condition (\SI{0.01}{\second}; Fig.\ref{fig:supplementary_fig_add}a), both methods yield comparable phase reconstructions, while PtyINR provides better object-amplitude recovery with fewer artifacts. At the higher-noise setting (\SI{0.001}{\second}; Fig.~\ref{fig:supplementary_fig_add}b), APG exhibits noisy object phase reconstructions, whereas PtyINR maintains high-quality phase estimates, as evidenced by the extracted patches. Although object amplitude reconstructions are suboptimal for both methods, PtyINR recovers a clearer object structure than APG. Moreover, PtyINR recovers a probe consistent with the \SI{0.01}{\second} result, whereas APG’s probe converges to a degenerate solution. 


\vspace{-10pt}
\subsection*{Conclusion}

In summary, the proposed PtyINR offers significant improvements over conventional approaches in both ideal and challenging experimental settings for X-ray ptychography reconstruction. By harnessing the expressive capacity of neural networks, PtyINR enables the recovery of illumination probes without relying on traditional iterative approaches and external measurement processes. This method enables accurate and robust recovery of both object and probe fields, even under low-dose and noise-limited conditions, establishing PtyINR as a highly promising framework for high-fidelity X-ray ptychographic imaging. Beyond X-ray ptychography, the flexibility of the PtyINR framework enables its extension to a variety of related imaging modalities, including near-field ptychography and advanced three-dimensional techniques such as Bragg ptychography. This adaptability underscores the potential of PtyINR to serve as a generalizable and powerful framework across a wide range of X-ray coherent imaging applications.

Although PtyINR requires slightly longer computational time compared with GPU-accelerated iterative methods due to its large number of trainable parameters, this cost is justified by its superior reconstruction quality and robustness. Currently, PtyINR has been enabled for multi-GPU acceleration, delivering improved efficiency. For example, reconstructing from diffraction patterns of size $40,401 \times 220 \times 220$ completed in 30 minutes using four RTX 3090 GPUs. Additionally, while this study assumes that probe positions are accurate, incorporating position correction algorithms~\cite{du2024predicting} into PtyINR presents another promising avenue for further development. Besides, though multi-mode probe reconstruction has been explored in the previous study \cite{thibault_reconstructing_2013}, this work adopts a single-mode probe assumption to focus on evaluating the core performance of PtyINR. Although fly-scan ptychography may introduce motion-induced blurring, PtyINR mitigates this effect by both leveraging the inherent deblurring capability of its object neural network and integrating a coherent-equivalent function into the estimated probe, thereby enhancing the accuracy of the final reconstruction. Future extensions of our method will include support for multi-mode probe modeling to further enhance reconstruction fidelity in more complex experimental conditions. The source codes with detailed instructions have been released at \href{https://github.com/TISGroup/PtyINR}{https://github.com/TISGroup/PtyINR}. 

\vspace{-10pt}
\section*{Methods}
\label{s:methods}

\subsection*{Forward Model}

Fig.~\ref{illustrate}a illustrates the experimental geometry of a typical X-ray ptychography setup. A coherent beam with focusing optics illuminates a sample in a pre-defined scanning fashion. Overlapping regions of illumination yield sets of far-field diffraction patterns that encode the exit wave generated by object $O$ and probe $P$, where only the intensity $I$ is recorded and the phase is lost. The forward process at the $j$th fly scanning position could be written as 
\begin{equation}
I_j(q) = \int_{t_0}^{t_0 + \Delta t} \left| \mathcal{F} \left\{ P(\mathbf{r} - \mathbf{v}t) \, O(\mathbf{r} + \mathbf{r}_j) \right\} \right|^2 dt,
\end{equation}
where $\mathbf{r}$ and $\mathbf{q}$ are the real-space and reciprocal-space coordinates, respectively. $\Delta t$ is the detector dwell time and $v$ is the scan speed~\cite{huang_fly-scan_2015}.

\vspace{-10pt}
\subsection*{The PtyINR Framework}

Incorporated the physical process into the training, PtyINR simulates the formation of diffraction patterns $I_p$ from predicted complex-valued object and probe functions. As shown in Fig.~\ref{illustrate}b, the difference between $I_p$ and the experimentally measured diffraction patterns $I_m$ is minimized using a hybrid loss function $\mathcal{L}$, and gradients are backpropagated to update the neural network weights. The developed loss function based on SmoothL1 $\hat{\mathcal{L}}$ \cite{girshick_fast_2015} can be explicitly written as:
\begin{align}
\mathcal{L} &= 
\begin{cases}
\hat{\mathcal{L}}(I_m, I_p,\beta) + \lambda  \bar{A_p}, & \text{if } t \leq k \\
\hat{\mathcal{L}}(I_m, I_p,\beta), & \text{if } t > k
\end{cases}, \quad
\hat{\mathcal{L}}(I_m, I_p,\beta) = 
\begin{cases}
\frac{1}{2\beta}(\sqrt{I_m} - \sqrt{I_p})^2, & \text{if } |\sqrt{I_m} - \sqrt{I_p}| < \beta \\
|\sqrt{I_m} - \sqrt{I_p}| - \frac{\beta}{2}, & \text{otherwise}.
\end{cases}
\end{align}

Here, $\bar{A_p}$ denotes the mean of the amplitude ${A_p}$ of the probe, which is controlled by the predefined coefficient $\lambda \geq 0$, and $t$ represents the current training step. $k$ is a hyperparameter that determines when to terminate the regularization during training, while $\beta$ controls the balance between mean squared error $\ell_2$ and $\ell_1$ loss, allowing the final loss to interpolate between the two. A more detailed discussion of the loss functions is provided in the \hyperref[Supplementary Note]{Supplementary Note}, Figs.~\ref{fig:supplementary_fig13} and~\ref{fig:supplementary_fig15}, where we present comprehensive ablation studies on various loss configurations.

The experimental object is decomposed into two components, amplitude and phase, each represented by an independent multilayer perceptron (MLP) conditioned on spatial coordinates (Fig.~\ref{illustrate}c). Both MLPs adopt the sinusoidal representation network (SIREN) architecture~\cite{sitzmann_implicit_2020}, which utilizes sine functions as activation units between layers. Each network consists of three hidden layers, with 512 units per layer. A sinusoidal activation function is applied after each linear transformation, following the form $\sin(\omega_i \mathbf{W} \mathbf{x} + \mathbf{b})$, where $\omega_i$ is a layer-specific frequency scaling factor (typically set to 30), $\mathbf{W}$ denotes the weight matrix, and $\mathbf{b}$ the bias vector. It is important to note that the frequency parameter in the first layer, $\omega_0$, plays a critical role in determining the network’s ability to represent high-frequency details in the object. Further details regarding the choice and impact of $\omega_0$ are provided in the \hyperref[Supplementary Note]{Supplementary Note}.

The network weights are initialized following the scheme proposed by Vincent et al.~\cite{sitzmann_implicit_2020}, which is specifically designed to facilitate the learning of high-frequency signals. The neural network takes the normalized 2D spatial coordinates of the object, scaled to the range [0, 1], as input and directly outputs the corresponding amplitude and phase of the object. The object amplitude $A_O$ and phase $\varphi_O$ are combined to form the complex-valued object function $O(r)$, where
\begin{equation}
O(\mathbf{r}) = \mathcal{A}_O(\mathbf{r}) \cdot \exp\left( i \cdot \mathcal{\varphi}_O(\mathbf{r}) \right).
\end{equation}
Simultaneously, the probe is also parameterized as two coordinate-based neural fields (Fig.~\ref{illustrate}d). Unlike the object, which is represented by SIREN, the probe adopts a multi-resolution hash encoding scheme, inspired by Instant-NGP (Fig.~\ref{illustrate}e) \cite{muller_instant_2022}, to efficiently encode spatial coordinates into learnable features. 

In this approach, the 2D input coordinate \( \mathbf{r} \in [0,1]^2 \) is scaled and discretized at multiple resolution levels \( \ell = 1, \dots, N \), where each level corresponds to a grid of increasing resolution, typically \( 2^\ell \times 2^\ell \). At each level, the coordinate is mapped to 4 neighboring grid points in 2D (the corners of its enclosing cell), and hashed indices are used to retrieve the corresponding feature vectors from a learnable table. These feature vectors are stored in compact hash tables to minimize memory usage. For each level, a bilinear interpolation is performed between the feature vectors based on the coordinate's relative position within the cell. The interpolated features from all levels are concatenated to form the final high-dimensional encoding \( \mathbf{f}(\mathbf{r}) \):
\begin{equation}
\mathbf{f}(\mathbf{r}) = \bigoplus_{\ell=1}^{N} \sum_{i=1}^{4} w_{\ell,i} \cdot \mathcal{H}_\ell(g_{\ell,i}),
\end{equation}

where \( \mathcal{H}_\ell(g_{\ell,i}) \) is the feature vector retrieved from the hash table at level \( \ell \) for the \( i \)-th neighbor grid point \( g_{\ell,i} \), and \( w_{\ell,i} \) is the bilinear interpolation weight based on the geometric distance to \( \mathbf{r} \).

This encoding scheme allows the network to efficiently capture both low- and high-frequency spatial features. The resulting feature vector \( \mathbf{f}(\mathbf{r}) \) is then passed through a shallow multilayer perceptron (MLP) to predict either the amplitude $A_P$ or phase $\varphi_P$ of the complex probe $P(r)$, where $P(r)$ can be expressed as:
\begin{equation}
P(\mathbf{r}) = \mathcal{A}_P(\mathbf{r}) \cdot \exp\left( i \cdot \mathcal{\varphi}_P(\mathbf{r}) \right)
\end{equation}
The architectural parameters of the probe neural networks follow the default configuration of Instant-NGP~\cite{muller_instant_2022}. Specifically, the resolution scale was set to 16, and each level of the multi-resolution hash grid utilized 2 learnable feature channels per entry. The maximum number of hash table entries per level was defined as $2^{15}$. The base resolution of the coarsest level was set to 16, with a per-level resolution growth factor fixed at 1.5. The outer ReLU-activated MLP is composed of two hidden layers, with each layer having 64 units. Although alternative parameter configurations were explored, this setup consistently yielded the best performance across all evaluated scenarios. A more detailed discussion on the model structures could be found in the \hyperref[Supplementary Note]{Supplementary Note}.

The entire PtyINR framework is implemented using a PyTorch backend~\cite{paszke_pytorch_2019}, with model optimization carried out via the Adam optimizer~\cite{kingma2015adam}, employing either full-batch gradient descent or mini-batch stochastic gradient descent. All simulation-based reconstructions were conducted on a single NVIDIA RTX 4090 GPU, while experimental data reconstruction was performed using an NVIDIA A100 SXM4 GPU. The learning rate for training the object and probe neural networks ranged from $10^{-5}$ to $10^{-4}$, while the weighting coefficient $\beta$ in the loss function varied between $10^{-4}$ and $10^{-1}$ depending on the specific task. Across all experiments, PtyINR maintained a consistent neural network architecture with a relatively small memory footprint. The complete framework comprises approximately 3.1 million trainable parameters, occupying around 20~MB of memory.

\vspace{-10pt}
\subsection*{Sample Preparation}

\subsubsection*{Simulation Experiments}

For the experimental object, we used a subset of the experimental X-ray ptychography dataset from Cherukara \textit{et al.} ~\cite{cherukara_ai-enabled_2020}, where a tungsten calibration sample was reconstructed using the ePIE algorithm. The full object has a resolution of $544 \times 544$ pixels. We cropped a $241 \times 241$ region and used only the central $181 \times 181$ pixels for evaluation to avoid boundary artifacts for conventional algorithms such as ePIE. The probe was obtained from a reconstruction performed during a real ptychography experiment conducted at Brookhaven National Laboratory and was cropped to its central $64 \times 64$ region.

To simulate the diffraction patterns, we set the energy of the incident probe to \SI{15}{\kilo\electronvolt} and the distance from the object to the detector to \SI{1.2}{\meter}, ensuring that the Fraunhofer far-field condition is satisfied. To focus solely on evaluating the recovery performance of different reconstruction algorithms, the simulation uses raster step scanning with precise positional accuracy to cover the entire sample.

In the experiments involving \textit{Poisson}, \textit{Gaussian}, and \textit{mixed} noise, noise is added to the simulated clean intensity $I_{\text{clean}}$ to obtain the noisy measurements $I_{\text{noisy}}$, defined as:
\begin{equation}
I_{\text{Poisson}} =\frac{\max(I_{\text{clean}})}{\alpha} \text{Poisson}\left( \frac{I_{\text{clean}}}{\max(I_{\text{clean}})} \cdot \alpha \right)
\end{equation}
\begin{equation}
I_{\text{Gaussian}} = I_{\text{clean}} + \mathcal{N}(0, \sigma^2)
\end{equation}
\begin{equation}
I_{\text{Mixed}} = I_{\text{Poisson}} + \mathcal{N}(0, \sigma^2)
\end{equation}

Here, $\alpha$ is a scaling factor representing the photon count level, and $\mathcal{N}(0, \sigma^2)$ denotes additive Gaussian noise with zero mean and variance $\sigma^2$. In our simulations, we set $\alpha = 10$ and $\sigma = 100$. All resulting negative intensity values are clipped to zero to ensure physical validity of the measurements.

\subsubsection*{Real Data Experiments on Scanning Step Sizes}

All real data experiments were conducted at the Hard X-ray Nanoprobe (HXN) beamline of the National Synchrotron Light Source II (NSLS-II) at Brookhaven National Laboratory. In the experiments conducted at varying scanning ratios using the FZP setup, the detector was positioned \SI{0.5}{\meter} downstream from the sample. The incident X-ray wavelength was set to \SI{0.12}{\nano\meter}, corresponding to a photon energy of approximately \SI{10}{\kilo\electronvolt}. Diffraction patterns were recorded using a pixel-array detector (Merlin, Quantum Detectors) with a pixel size of \SI{55}{\micro\meter} with fly-scan mode.

The sample used in FZP was a single-crystalline battery cathode particle LiCo$\text{O}_2$. Raster fly-scanning was performed over a total scan area of \SI{5.47}{\micro\meter} $\times$ \SI{5.27}{\micro\meter}. Each diffraction pattern was recorded with a resolution of $112 \times 112$ pixels. For a scanning step size of \SI{30}{\nano\meter} in each direction, a total of 19,600 scan positions were acquired. When the step size was increased to \SI{60}{\nano\meter}, the number of scan positions was reduced to 4,900. At a step size of \SI{120}{\nano\meter}, there were 1,225 scan positions collected. 

For the experiment conducted using the MLL focusing system, the incident X-ray wavelength was set to \SI{0.08}{\nano\meter}, corresponding to a photon energy of approximately \SI{15}{\kilo\electronvolt}. The detector was positioned \SI{1.055}{\meter} downstream from the sample and featured a pixel size of \SI{75}{\micro\meter}. The sample used in this experiment was a Siemens star test pattern fabricated from gold (Au). A raster fly-scan scheme was employed over an area of \SI{3.17}{\micro\meter} $\times$ \SI{3.16}{\micro\meter}, and diffraction patterns were recorded at a resolution of $220 \times 220$ pixels. For a scanning step size of \SI{10}{\nano\meter} $\times$ \SI{10}{\nano\meter}, a total of 40,401 scan positions were acquired. At a step size of \SI{30}{\nano\meter} $\times$ \SI{30}{\nano\meter}, 4,489 scan positions were collected, and for a step size of \SI{40}{\nano\meter} $\times$ \SI{40}{\nano\meter}, 2,525 scan positions were recorded.

\subsubsection*{Real Data Experiments on Exposure Times}

Experiments were performed on a Siemens-star test pattern fabricated from gold (Au), distinct from that used in the step-size variation study. Probes were generated using the MLL system. For the contrast dataset with exposure times of \SI{0.2}{\second} and \SI{0.003}{\second}, the incident X-ray wavelength was \SI{0.08}{\nano\meter} (corresponding to a photon energy of approximately \SI{15}{\kilo\electronvolt}). The detector, with a pixel size of \SI{75}{\micro\meter}, was positioned \SI{1.055}{\meter} downstream of the sample. For the dataset with exposure times of \SI{0.01}{\second} and \SI{0.001}{\second}, the wavelength was \SI{0.1}{\nano\meter}; the detector (pixel size \SI{75}{\micro\meter}) was positioned \SI{1}{\meter} downstream.

For the \SI{0.2}{\second}/\SI{0.003}{\second} dataset, the Siemens star was fly-scanned on a raster grid covering \SI{3.16}{\micro\meter} $\times$ \SI{3.16}{\micro\meter}, with step sizes of \SI{10}{\nano\meter} in both directions, yielding $201 \times 201$ diffraction patterns. For the \SI{0.1}{\second}/\SI{0.001}{\second} dataset, the raster scan covered \SI{3.12}{\micro\meter} $\times$ \SI{3.12}{\micro\meter}, with step sizes of \SI{30}{\nano\meter} $\times$ \SI{20}{\nano\meter}, yielding $100 \times 100$ diffraction patterns.

\subsection*{Evaluation Metrics}

To quantitatively assess reconstruction quality, we adopted various evaluation metrics for simulated and experimental data. For simulated datasets, where ground truth is available, we employed the PSNR as the primary metric. PSNR measures the fidelity of the reconstructed object with respect to the ground truth and is defined as:
\begin{equation}
\mathrm{PSNR} = 10 \cdot \log_{10} \left( \frac{\mathrm{MAX}^2}{\mathrm{MSE}} \right),
\end{equation}

where $\mathrm{MAX}$ denotes the maximum possible pixel value of the image and $\mathrm{MSE}$ is the mean squared error between the reconstructed image and the ground truth.

Due to the inherent global phase ambiguity in ptychographic reconstructions, the reconstructed complex object often contains an arbitrary global phase shift. To accurately compare the reconstructed phase with the ground truth, we corrected this global phase offset by minimizing the angular difference between the two. Specifically, we estimated the optimal global phase shift $\theta$ that minimizes the discrepancy between the phase of the ground truth $O$ and the phase of the reconstructed object $\hat{O}$, adjusted by $e^{i\theta}$:
\begin{equation}
\theta^{*} = \arg\min_{\theta} \left\| \angle(O) - \angle\left( \hat{O} \cdot e^{i\theta} \right) \right\|_2^2,
\end{equation}

where the $\ell_2$ norm is computed over all pixels. Once the optimal phase shift $\theta^*$ is found, the reconstruction is aligned by multiplying $\hat{O}$ with $e^{i\theta^*}$ before computing quantitative metrics such as PSNR.

For experimental datasets, where ground truth is not accessible, we utilized the Fourier Ring Correlation (FRC) \cite{koho_fourier_2019} to estimate the spatial resolution of the reconstructions. The FRC quantifies the correlation between two independently reconstructed images in the Fourier domain, and is computed as:
\begin{equation}
\mathrm{FRC}(f) = \frac{\sum_{k \in R(f)} F_1(k) \cdot F_2^*(k)}{\sqrt{\sum_{k \in R(f)} |F_1(k)|^2 \cdot \sum_{k \in R(f)} |F_2(k)|^2}},
\end{equation}

where $F_1(k)$ and $F_2(k)$ are the Fourier transforms of two independent reconstructions, $F_2^*(k)$ is the complex conjugate of $F_2(k)$, and the summation is carried out over the ring $R(f)$ at the spatial frequency $f$. The resolution is determined based on the spatial frequency at which the FRC curve falls below a predefined threshold. In this work, the threshold was defined using the half-bit criterion.

\section*{Author Contributions}
\label{s:contributions}

J. Li and X. Huang conceptualized the study. T. Li and Z. Xu implemented the reconstruction model. Experimental data collection was carried out by Z. Gao, H. Yan, and X. Huang. The manuscript was written by T. Li, X. Huang, and J. Li, with valuable feedback and input from all authors. 

\section*{Acknowledgements}
\label{s:ackno}
This work at The Chinese University of Hong Kong (CUHK) is supported by the National Natural Science Foundation of China (Nos. T2422017 and 52303301), the Hong Kong Research Grants Council (No. 21204124), and the Shun Hing Institute of Advanced Engineering, CUHK (No. RNE-p1-25). This research used the Hard X-ray Nanoprobe beamline at 3-ID of the National Synchrotron Light Source II (NSLS-II), a US Department of Energy (DOE) Office of Science User Facility operated for the DOE Office of Science by Brookhaven National Laboratory under contract no. DE-SC0012704.


\onecolumn 
\fancyhead{} 
\renewcommand{\floatpagefraction}{0.1}
\lfoot[\bSupInf]{\dAuthor}
\rfoot[\dAuthor]{\cSupInf}
\newpage

\captionsetup*{format=largeformat} 
\setcounter{figure}{0} 
\setcounter{equation}{0} 
\makeatletter 
\renewcommand{\thefigure}{S\@arabic\c@figure} 
\makeatother
\def\theequation{S\arabic{equation}}


\newpage
\section*{Supplementary Information}

\hspace{-6em}\textbf{Supplementary Note}
{
\setlength{\parindent}{2em} 
\setlength{\leftskip}{6em}  
\setlength{\parindent}{-5.1em} 

\hspace{-1em}\textbf{Figure S1.} \hspace{10pt}Illustrations of probe size estimation and overlap scenarios.

\hspace{-1em}\textbf{Figure S2.} \hspace{10pt}Probe-known object reconstruction under varying overlap ratios.  

\hspace{-1em}\textbf{Figure S3.} \hspace{10pt}Object reconstruction with unknown probes.  

\hspace{-1em}\textbf{Figure S4.} \hspace{10pt}Probe reconstruction with unknown probes.  

\hspace{-1em}\textbf{Figure S5.} \hspace{10pt}Comparison of object reconstruction under different noise conditions.  

\hspace{-1em}\textbf{Figure S6.} \hspace{10pt}Comparison of probe reconstruction under different noise conditions.  

\hspace{-1em}\textbf{Figure S7.} \hspace{10pt}Reconstructed object phase for the battery sample dataset acquired with varying scanning step sizes.  

\hspace{-1em}\textbf{Figure S8.} \hspace{3pt}Reconstructed probe amplitude and phase for the battery sample dataset acquired with varying \hspace{20pt}scanning step sizes.  

\hspace{-1em}\textbf{Figure S9.} \hspace{10pt}Reconstructed object amplitude and phase corresponding to the Siemens-star test pattern dataset.  

\hspace{-1em}\textbf{Figure S10.} \hspace{5pt}Additional comparisons with reconstructed object amplitude and phase corresponding to the Siemens star test pattern dataset.  

\hspace{-1em}\textbf{Figure S11.} \hspace{5pt}Reconstructed probe functions corresponding to the Siemens-star test pattern dataset.  

\hspace{-1em}\textbf{Figure S12.} \hspace{5pt}Reconstruction results of the Siemens-star test pattern at an exposure time of 0.2 seconds.  

\hspace{-1em}\textbf{Figure S13.} Reconstruction results of the Siemens-star test pattern using RAAR and AD methods at different exposure times.  

\hspace{-1em}\textbf{Figure S14.} Reconstruction results of the Siemens-star test pattern using APG and PtyINR methods at 0.01s and 0.001s exposure times.  

\hspace{-1em}\textbf{Figure S15.} \hspace{5pt}Probe initialization and ablation study of probe recovery procedures for PtyINR.  

\hspace{-1em}\textbf{Figure S16.} The PSNR comparison of different modules in PtyINR for object and probe recovery under varying overlap ratios.  

\hspace{-1em}\textbf{Figure S17.} \hspace{5pt}Evaluation of loss functions and first omega parameter in PtyINR.  

}
\clearpage
\section*{Supplementary Note}
\label{Supplementary Note}
\subsection*{Loss Function}
It is noteworthy that, rather than directly applying the mean squared error $\ell_2$ between $I_p$ and $I_m$ as in prior works, we systematically investigated the effect of different loss functions on the probe recovery performance. As shown in Fig.~\ref{fig:supplementary_fig15}a, we observed that in the presence of Poisson noise, $\ell_1$ loss tends to oversmooth fine details, while $\ell_2$ better preserves both contrast and structural fidelity. However, Fig.~\ref{fig:supplementary_fig15}b also reveals that $\ell_1$ loss generally leads to more stable convergence during probe recovery, whereas $\ell_2$ often suffers from abrupt instabilities and training crashes, especially when the probe becomes quite complicated. To balance these trade-offs, we adopted the SmoothL1 $\hat{\mathcal{L}}$ \cite{girshick_fast_2015} loss function, which interpolates between $\ell_1$ and $\ell_2$, for both object and probe reconstruction tasks.

Furthermore, PtyINR employs random initialization for both the object and the probe, enabling blind recovery without imposing prior assumptions. While this design allows flexibility, it also introduces challenges due to the mutual dependency between the object and the probe during optimization. In particular, as demonstrated in Fig.~\ref{fig:supplementary_fig13}, direct reconstruction from random initialization can fail when the probe is from a Fresnel zone plate, causing the probe to diverge rather than converge into a focused solution.

To ensure stable convergence, we introduce a probe regularization term, $\lambda \bar{A_p} $, where the mean of the amplitude of the probe is multiplied with $\lambda\geq0$, applied during the initial training steps ($t \leq k$), along with a normalization constraint given by $P \leftarrow \frac{P}{\max(A_p)}$ throughout the whole training. These strategies guide the probe towards a physically plausible solution in stages of training. The effectiveness of this regularization and normalization scheme is further validated in Fig.~\ref{fig:supplementary_fig13} across different probe types.


\subsection*{Model Architectures}

Though it may appear reasonable to assign identical neural network architectures to both the object and probe—given their mathematical duality in the ptychographic reconstruction process, their inherent characteristics differ substantially. Each diffraction pattern encodes the interference between the full probe and only a localized region of the object. As a result, the probe contributes globally to every diffraction pattern, whereas each pattern contains information from only a small portion of the object. Consequently, a minor change in the recovered probe can significantly affect the entire reconstructed object. In contrast, a localized change in the object influences only a small subset of the training data and, therefore, has a limited impact on the overall probe recovery. Thus, during reconstruction, it is crucial to ensure that the probe is recovered in a stable manner, minimizing the risk of unexpected training instabilities or convergence failures. Furthermore, in practical imaging scenarios, objects typically contain fewer high-frequency components than probes. Objects tend to exhibit smooth intensity and phase variations, especially around edges, whereas probes—shaped by complex focusing optics—often include sharp phase transitions and high-frequency structures. These differences are visually evident in Figs.~\ref{fig:supplementary_fig6}, ~\ref{fig:supplementary_fig7}, ~\ref{fig:supplementary_fig9}, and ~\ref{fig:supplementary_fig10}, where the object exhibits gradual spatial variations, in contrast to the probe, which contains sharper edge features and finer-scale details.

Based on the distinct characteristics of probes and objects, we considered two types of implicit neural representations as backbone architectures: sine-activated networks (SIREN) \cite{sitzmann_implicit_2020} and multi-resolution ReLU-activated networks (ReLU) \cite{muller_instant_2022}. Both architectures are capable of modeling high-frequency details. In our design, we selected SIREN as the backbone to represent the object and ReLU for the probe representation. The choice of SIREN for the object network is motivated by prior findings showing that sine-activated networks are particularly effective at reconstructing sharp edges and linear features~\cite{sitzmann_implicit_2020}. These properties suggest that SIREN is well-suited for modeling object structures, which typically exhibit smooth intensity and phase variations, while also providing robustness against edge ringing and other possible artifacts introduced by noise or insufficient data. The results presented in Fig.~\ref{fig:supplementary_fig1} demonstrate that SIREN significantly outperforms a matrix-based representation (AD) for modeling the object when the probe is known. This comparison effectively serves as an ablation study, highlighting the advantages of using SIREN for object representation. 

Additionally, we investigated the feasibility of employing multi-resolution ReLU-activated MLPs~\cite{muller_instant_2022} for object modeling by replacing the SIREN backbone with the probe network architecture used in PtyINR. However, in scenarios where the probe was fixed and only the object was reconstructed, the sine-activated MLP consistently outperformed its ReLU-based counterpart, particularly under conditions of reduced overlap ratio, as illustrated in Fig.~\ref{fig:supplementary_fig14}. This performance disparity underscores the advantages of sine activations, whose oscillatory nature and spectral bias make them especially well-suited for capturing the smooth yet structurally rich variations typically present in object distributions.

Despite its expressive capacity, SIREN is susceptible to training instabilities. In our experiments, we explored the use of SIREN to represent both the object and the probe. Nevertheless, in experimental settings where one of the components—either the object or the probe—was known while the other remained unknown to be reconstructed, we observed that the reconstructions occasionally exhibited sudden degradations in quality during training epochs, which later recovered spontaneously without external intervention. This behavior was consistently observed. While such temporary degeneration appears to have no lasting effect on the final reconstruction, since the model typically recovers during training, this is not the case when both the object and the probe are simultaneously unknown. Under this condition, the reconstruction may fail entirely with a non-negligible probability, often resulting in complete and irreversible degradation. This observation reinforces the importance of ensuring stable probe recovery, as previously discussed. We attribute these instabilities to the oscillatory nature of the sine activation function employed in SIREN, which can lead to heightened sensitivity in gradient propagation and, consequently, to convergence challenges during optimization.

Therefore, to better parameterize the probe, we employed the multi-resolution ReLU-activated neural network architecture, inspired by the framework Instant-ngp~\cite{muller_instant_2022}. This architecture is capable of capturing fine-scale details with rapid convergence. More importantly, our empirical results demonstrate that these ReLU-based networks exhibit stable training behavior in probe reconstructions and do not suffer from the reconstruction degradation observed before. These characteristics make the ReLU-activated network particularly well-suited for modeling probes, which often contain high-frequency features and sharp edge variations.

To validate the effectiveness of our architecture, we conducted ablation studies using simulated datasets to compare our design against alternative parameterizations. Specifically, we replaced the probe neural network with two directly parameterized tensors, following a design similar to the auto-differentiation (AD) approach~\cite{du_adorym_2021}. As illustrated in Fig.~\ref{fig:supplementary_fig14}, when the object was fixed and only the probe was reconstructed, this alternative yielded significantly inferior performance relative to our default architecture. In contrast, the ReLU-activated MLP achieved superior results, particularly under low overlap conditions. These findings can be attributed to the hierarchical structure of the multi-resolution ReLU network, which facilitates faster learning and better generalization compared to directly parameterized matrices. By leveraging the expressive capacity of neural networks and a multi-resolution scheme, our approach enables the extraction of more meaningful features from the measured data and achieves more accurate reconstructions.

\subsection*{Important hyperparameter, the first omega ($\omega_0$)}

In addition, we investigated the effect of a critical hyperparameter in our object neural network: the first-layer frequency scaling factor, commonly referred to as the \textit{first omega} ($\omega_0$). This parameter is applied exclusively to the first layer of the network and determines the frequency scale of the input features. By modulating the periodicity of the sine activation functions, $\omega_0$ directly influences the representational capacity and learning behavior of the network.

Notably, this parameter offers a mechanism to control the model’s sensitivity to data sufficiency and noise. When training data is abundant and clean, a higher value of $\omega_0$ (e.g., 90) enables the network to capture finer details, resulting in improved reconstruction quality. This is evident in Fig.~\ref{fig:supplementary_fig15}, where for scanning step sizes of 30~nm and 60~nm, higher values of $\omega_0$ outperform lower ones (e.g., 30). However, excessively high values of $\omega_0$, such as 300, tend to induce periodic artifacts and degrade reconstruction quality.

Conversely, when the training data is limited or noisy, lower values of $\omega_0$ (e.g., around 30) yield more robust results by suppressing high-frequency artifacts. As shown in the bottom row of Fig.~\ref{fig:supplementary_fig15}, with a scanning step size of 120~nm---where the data is significantly undersampled---higher values of $\omega_0$ (90 - 300) introduce strong artifacts, whereas lower values produce reasonable reconstructions. Overall, the choice of $\omega_0$ provides a flexible mechanism to adjust the effective resolution scale of the reconstruction according to the availability and quality of the training data.

\clearpage

\section*{Supplementary Figure 1}

\begin{figure}[ht]
    \centering
    \includegraphics[width=\textwidth, height=\textheight, keepaspectratio]{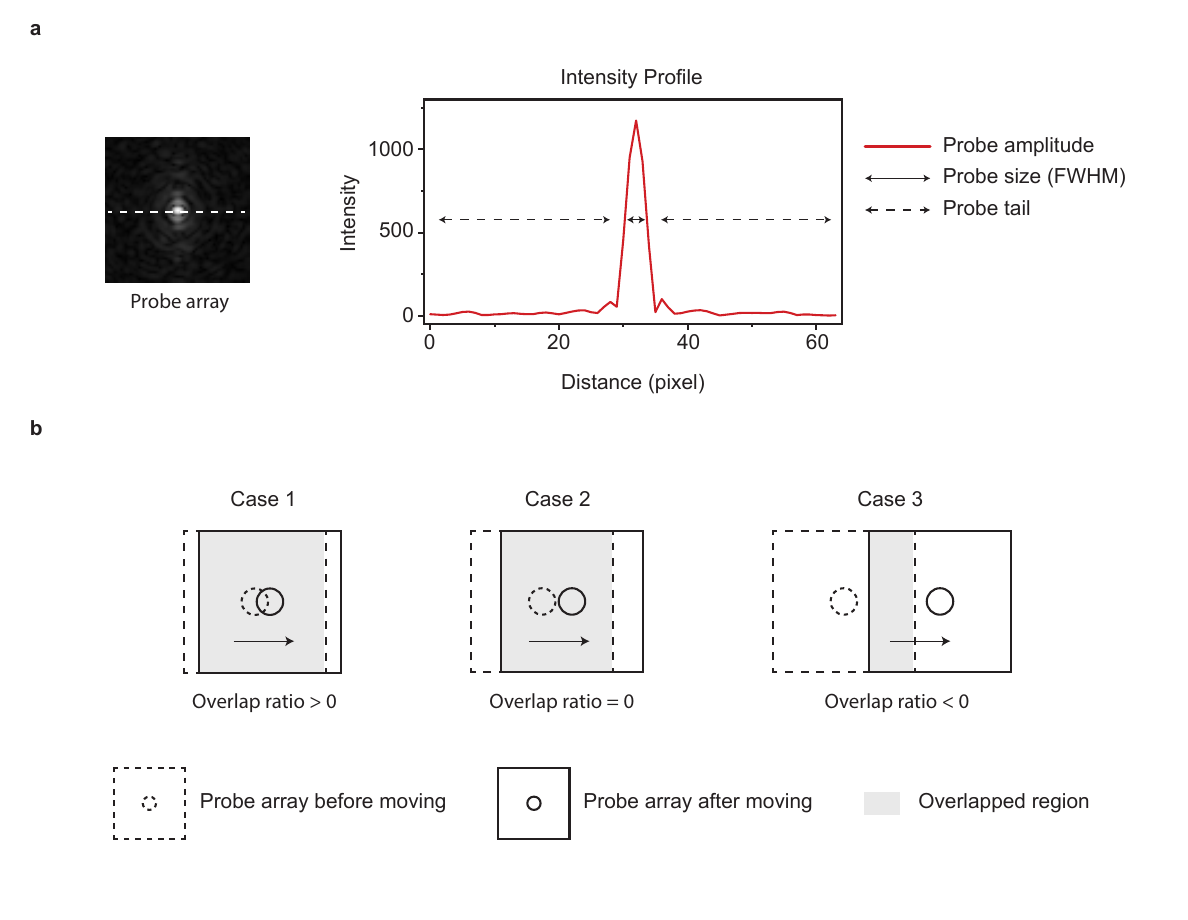}
    \caption{\textbf{Illustrations of probe size estimation and overlap scenarios} 
    (\textbf{a}) Probe array used in simulation experiments, along with the method for determining probe size by analyzing the intensity profile of the probe amplitude using the full width at half maximum (FWHM).
    (\textbf{b}) Illustration of probe overlap scenarios with positive, zero, and negative overlap ratios, highlighting that even in the case of a negative overlap ratio, certain regions of the scanning area remain overlapped.}
    \label{fig:supplementary_fig0}
\end{figure}
\clearpage

\section*{Supplementary Figure 2}

\begin{figure}[ht]
    \centering
    \includegraphics[width=0.9\textwidth, height=\textheight, keepaspectratio]{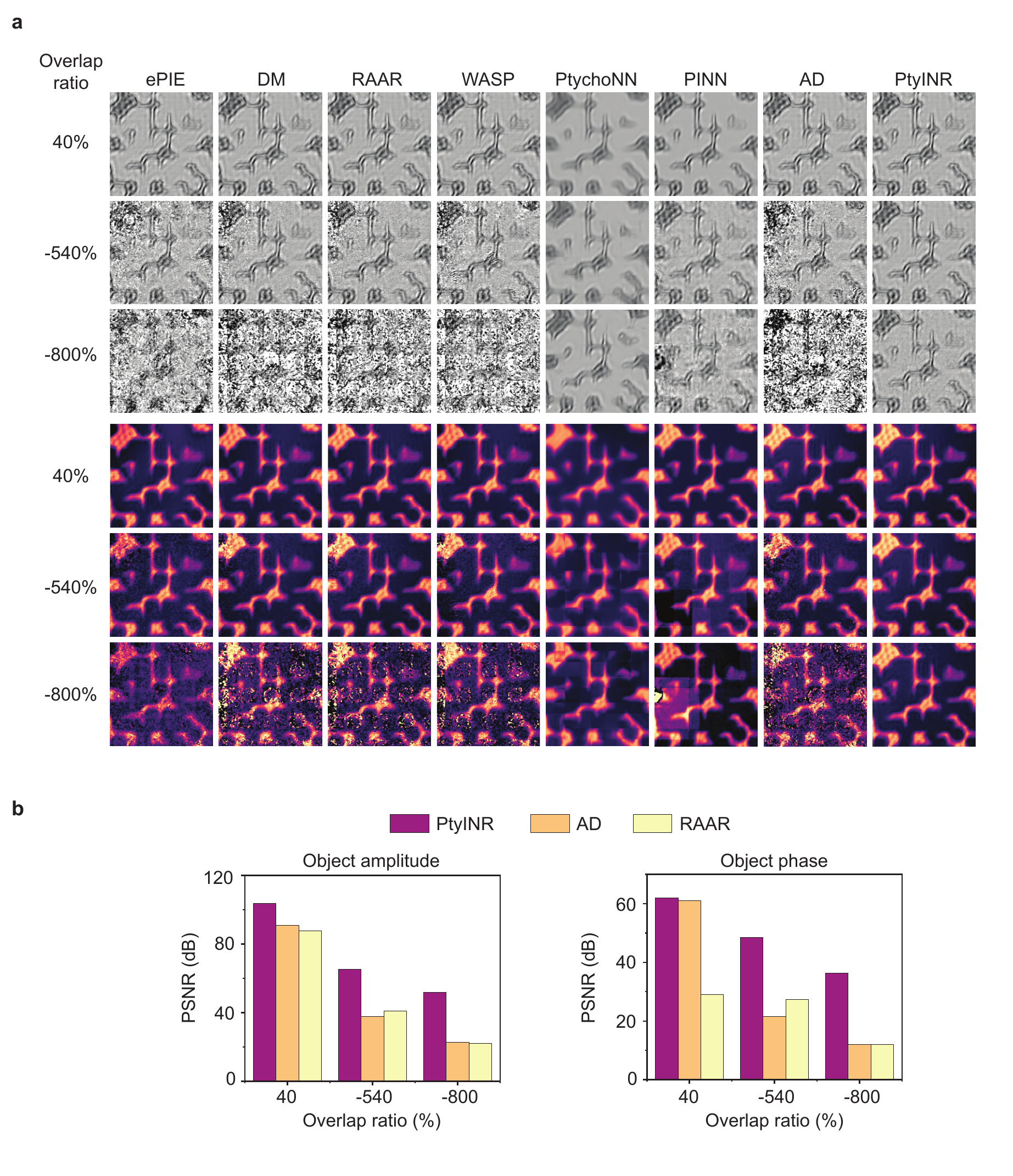}
    \caption{\textbf{Probe-known object reconstruction under varying overlap ratios.} 
    (\textbf{a}) Reconstruction results for object \textit{amplitude} (top three rows) and \textit{phase} (bottom three rows) using various algorithms under overlap ratios of 40\%, -540\%, and -800\%. The methods compared include ePIE, DM, RAAR, WASP, PtychoNN, PINN, AD, and PtyINR. 
    (\textbf{b}) Comparison of reconstruction accuracy (PSNR) between PtyINR, AD, and RAAR algorithms across varying probe overlap ratios. Quantitative results for other phase retrieval methods are shown in Fig.~\ref{simulate_overlap}.}
    \label{fig:supplementary_fig1}
\end{figure}
\clearpage

\section*{Supplementary Figure 3}
\begin{figure}[ht]
    \centering
    \includegraphics[width=0.9\textwidth, height=0.9\textheight, keepaspectratio]{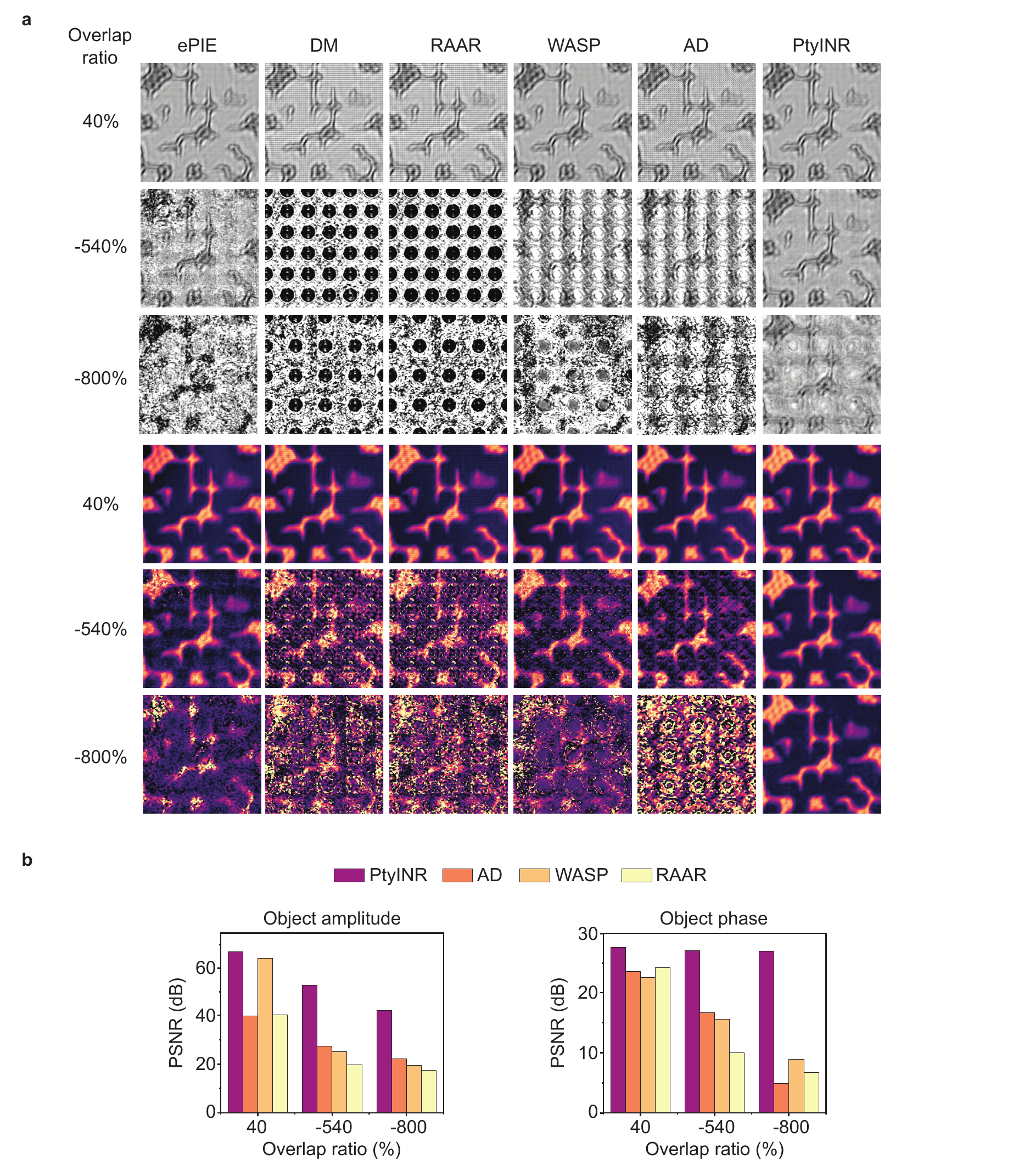}
    \caption{\textbf{Object reconstruction with unknown probes.} 
    (\textbf{a}) Reconstruction results for the object \textit{amplitude} (top three rows) and \textit{phase} (bottom three rows) under different overlap ratios: 40\%, -540\%, and -800\%. The compared methods include ePIE, DM, RAAR, WASP, AD, and PtyINR. 
    (\textbf{b}) Comparison of reconstruction accuracy (PSNR) between PtyINR, AD, WASP, and RAAR algorithms across varying probe overlap ratios. Quantitative results for other phase retrieval methods are shown in Fig.~\ref{simulate_overlap}.}
    \label{fig:supplementary_fig2}
\end{figure}
\clearpage

\section*{Supplementary Figure 4}
\begin{figure}[ht]
    \centering
    \includegraphics[width=0.9\textwidth, height=0.9\textheight, keepaspectratio]{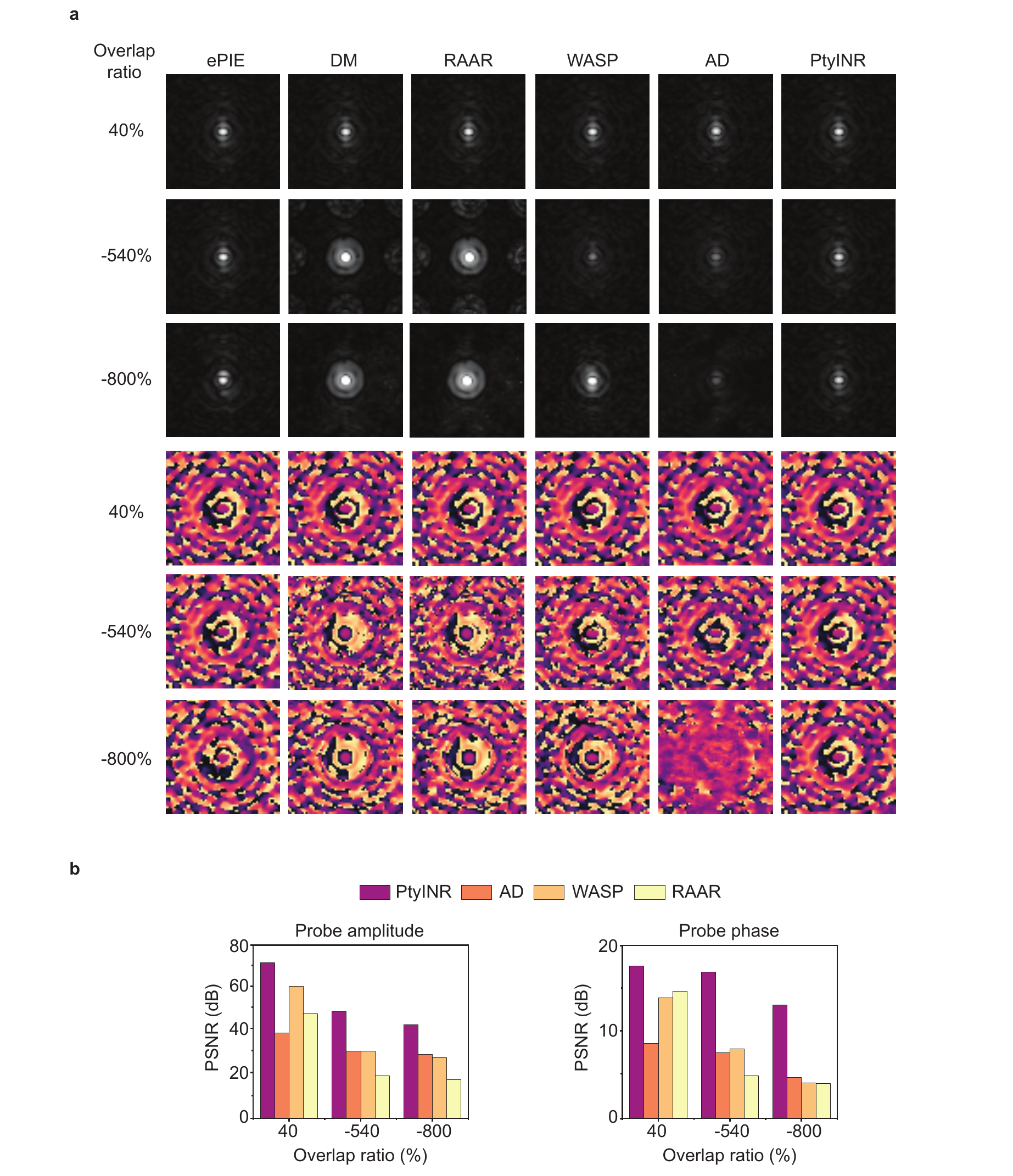}
    \caption{\textbf{Probe reconstruction with unknown probes.}
    (\textbf{a}) Recovered probe \textit{amplitude} (top three rows) and \textit{phase} (bottom three rows) using different methods under varying overlap ratios (40\%, -540\%, and -800\%). The methods include ePIE, DM, RAAR, WASP, AD, and PtyINR.
    (\textbf{b}) Comparison of reconstruction accuracy (PSNR) between PtyINR, AD, WASP, and RAAR algorithms across varying probe overlap ratios. Quantitative results for other phase retrieval methods are shown in Fig.~\ref{simulate_overlap}.
    }
    \label{fig:supplementary_fig3}
\end{figure}
\clearpage

\section*{Supplementary Figure 5}
\begin{figure}[ht]
    \centering
    \includegraphics[width=0.95\textwidth, height=0.95\textheight, keepaspectratio]{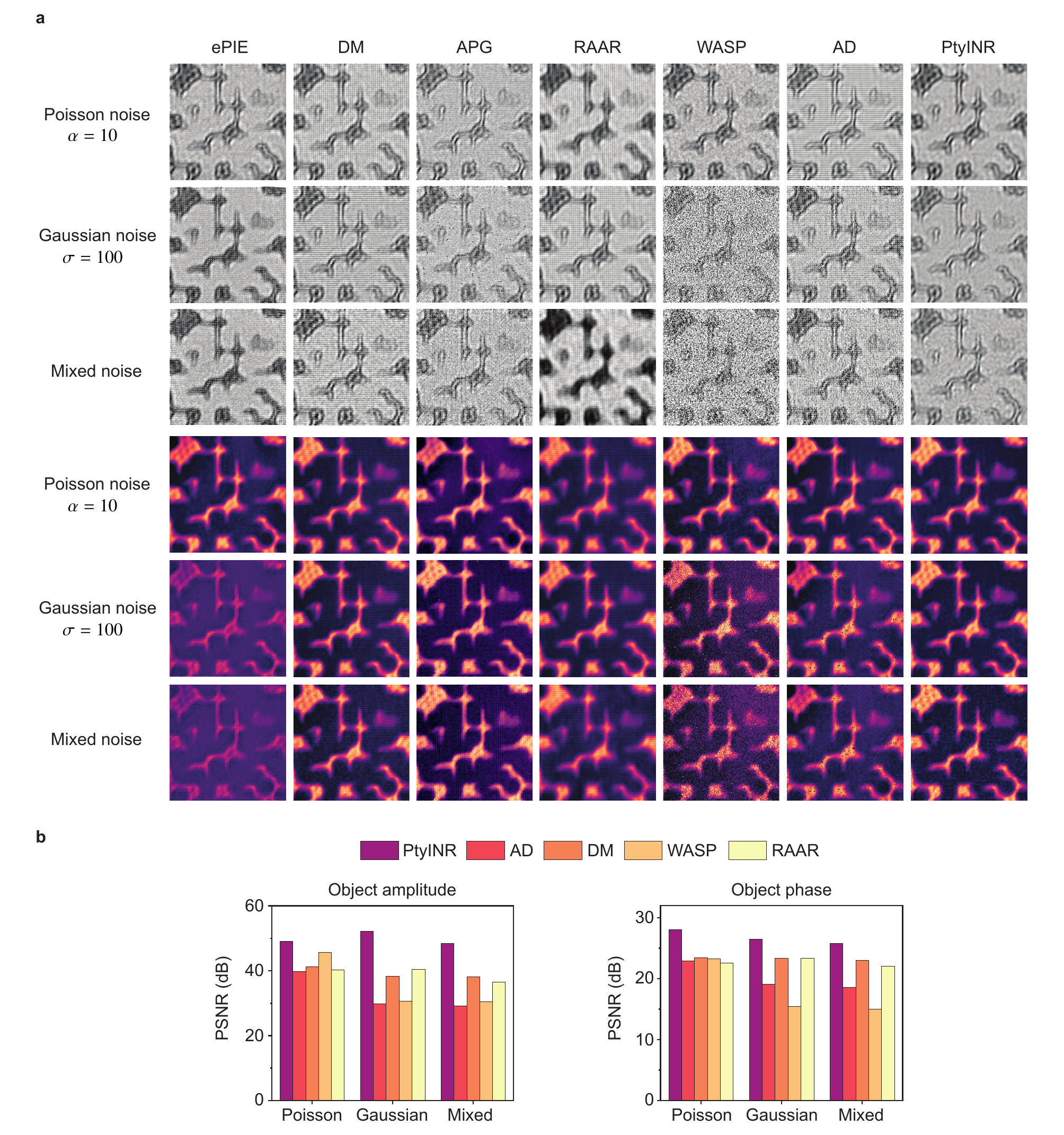}
    \caption{\textbf{Comparison of object reconstruction under different noise conditions.}
    (\textbf{a}) Reconstruction results of the object \textit{amplitude} (top three rows) and \textit{phase} (bottom three rows) using different methods (ePIE, DM, APG, RAAR, WASP, AD, and PtyINR) under Poisson noise, Gaussian noise, and combined noise (Poisson + Gaussian).
    (\textbf{b}) Comparison of reconstruction accuracy (PSNR) between PtyINR, AD, DM, WASP, and RAAR algorithms across three noise conditions. Quantitative results for other phase retrieval methods are shown in Fig.~\ref{simulate_noise}.
    }
    \label{fig:supplementary_fig4}
\end{figure}
\clearpage

\section*{Supplementary Figure 6}
\begin{figure}[ht]
    \centering
    \includegraphics[width=0.95\textwidth, height=0.95\textheight, keepaspectratio]{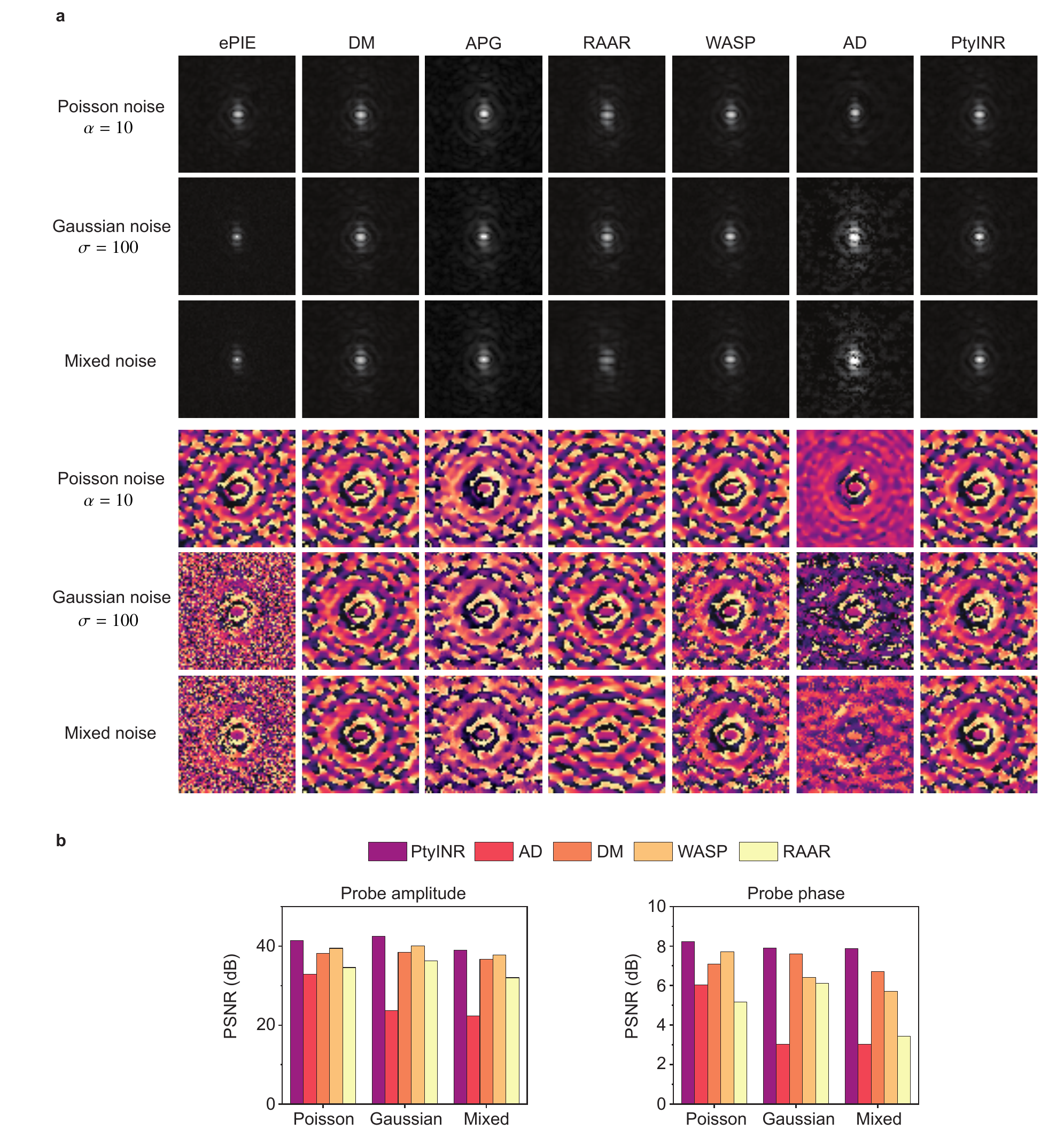}
    \caption{\textbf{Comparison of probe reconstruction under different noise conditions.}
    (\textbf{a}) Reconstructed probe \textit{amplitude} (top three rows) and \textit{phase} (bottom three rows) using various methods (ePIE, DM, APG, RAAR, WASP, AD, and PtyINR) under Poisson noise, Gaussian noise, and combined noise (Poisson + Gaussian).
    (\textbf{b}) PSNR comparison between the reconstructed probes and the ground truth under each noise condition, quantifying the reconstruction accuracy. Quantitative results for other phase retrieval methods are shown in Fig.~\ref{simulate_noise}.
    }
    \label{fig:supplementary_fig5}
\end{figure}
\clearpage

\section*{Supplementary Figure 7}
\begin{figure}[ht]
    \centering
    \includegraphics[width=\textwidth, height=\textheight, keepaspectratio]{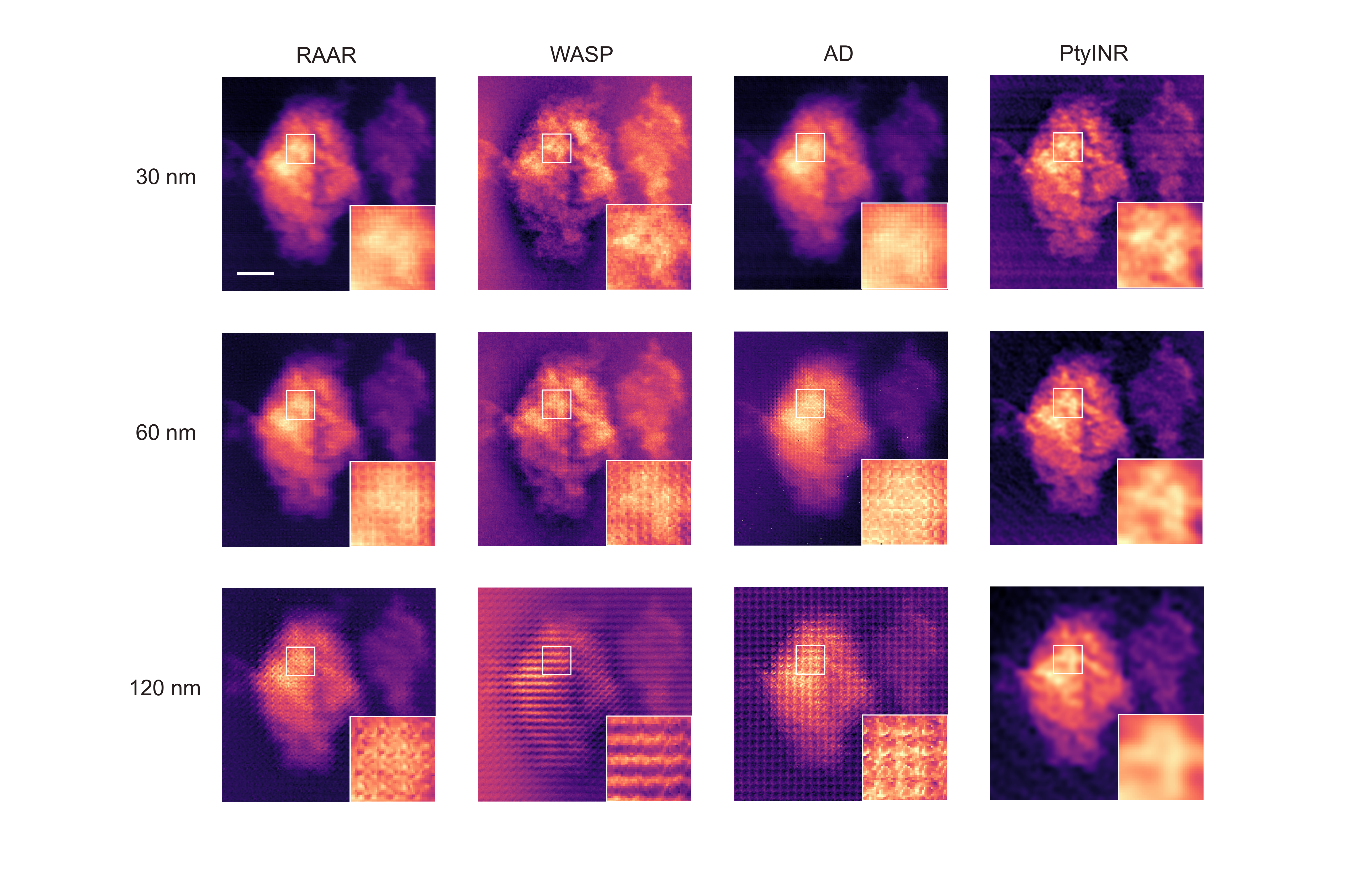}
    \caption{\textbf{Reconstructed object phase for the battery sample dataset acquired with varying scanning step sizes.}
    Reconstructed phase images using different methods (RAAR, WASP, AD, and PtyINR) for scanning step sizes of 30 nm, 60 nm, and 120 nm. The units of the reconstructed phase values are in radians. Insets display magnified regions of interest with a 3× zoom factor. Scale bar: 600 nm.
    }
    \label{fig:supplementary_fig6}
\end{figure}
\clearpage

\section*{Supplementary Figure 8}
\begin{figure}[ht]
    \centering
    \includegraphics[width=0.9\textwidth, height=0.9\textheight, keepaspectratio]{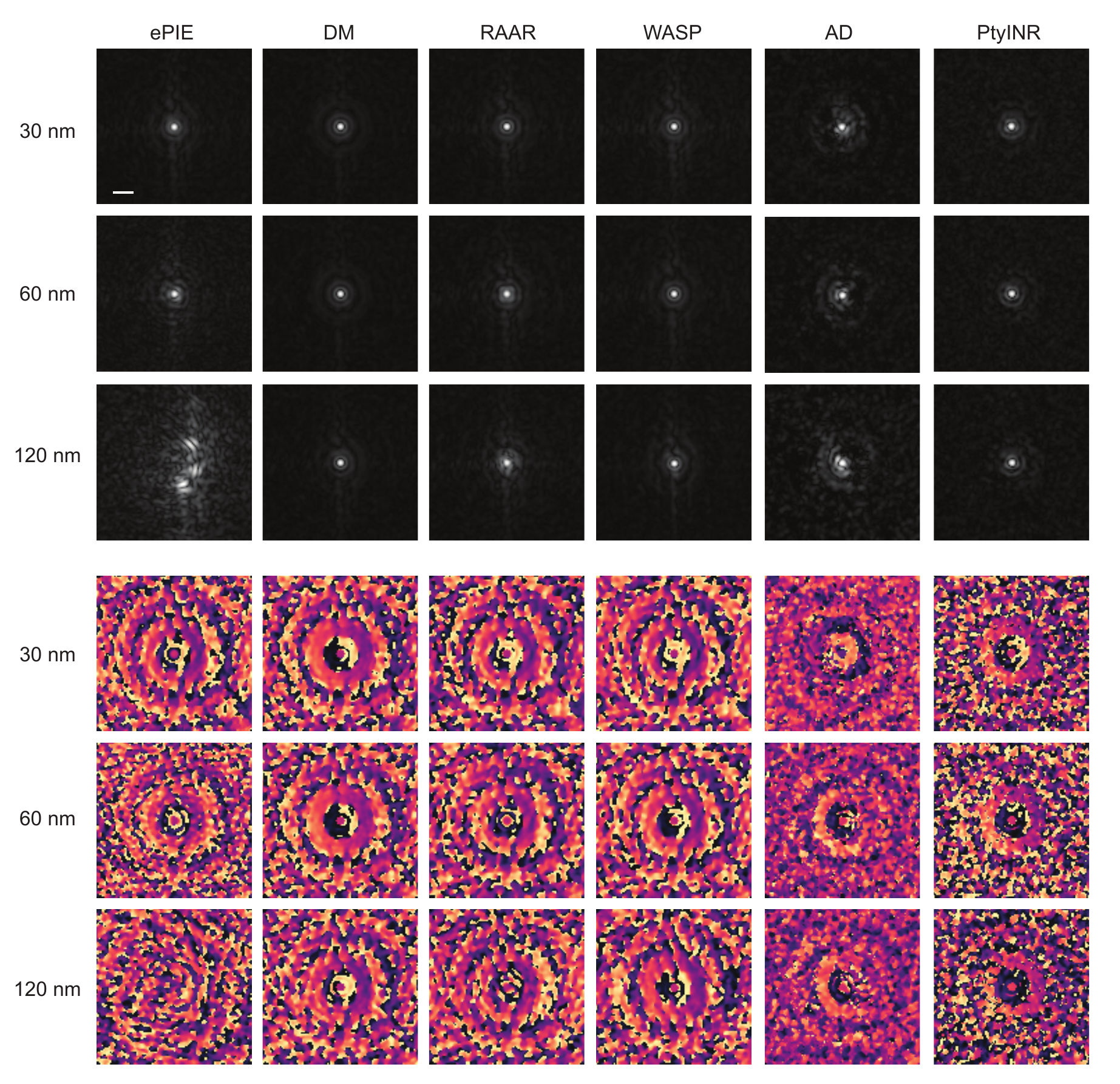}
    \caption{\textbf{Reconstructed probe amplitude and phase for the battery sample dataset acquired with varying scanning step sizes.  }
    Probe reconstruction results using different methods (ePIE, DM, RAAR, WASP, AD, and PtyINR) at scanning step sizes of 30 nm, 60 nm, and 120 nm. For each method and step size, both the probe amplitude (top three rows) and phase (bottom three rows) are visualized to assess the robustness and fidelity of probe recovery under varying spatial sampling conditions. Scale bar: 150 nm.
    }
    \label{fig:supplementary_fig7}
\end{figure}
\clearpage

\section*{Supplementary Figure 9}
\begin{figure}[ht]
    \centering
    \includegraphics[width=0.85\textwidth, height=0.85\textheight, keepaspectratio]{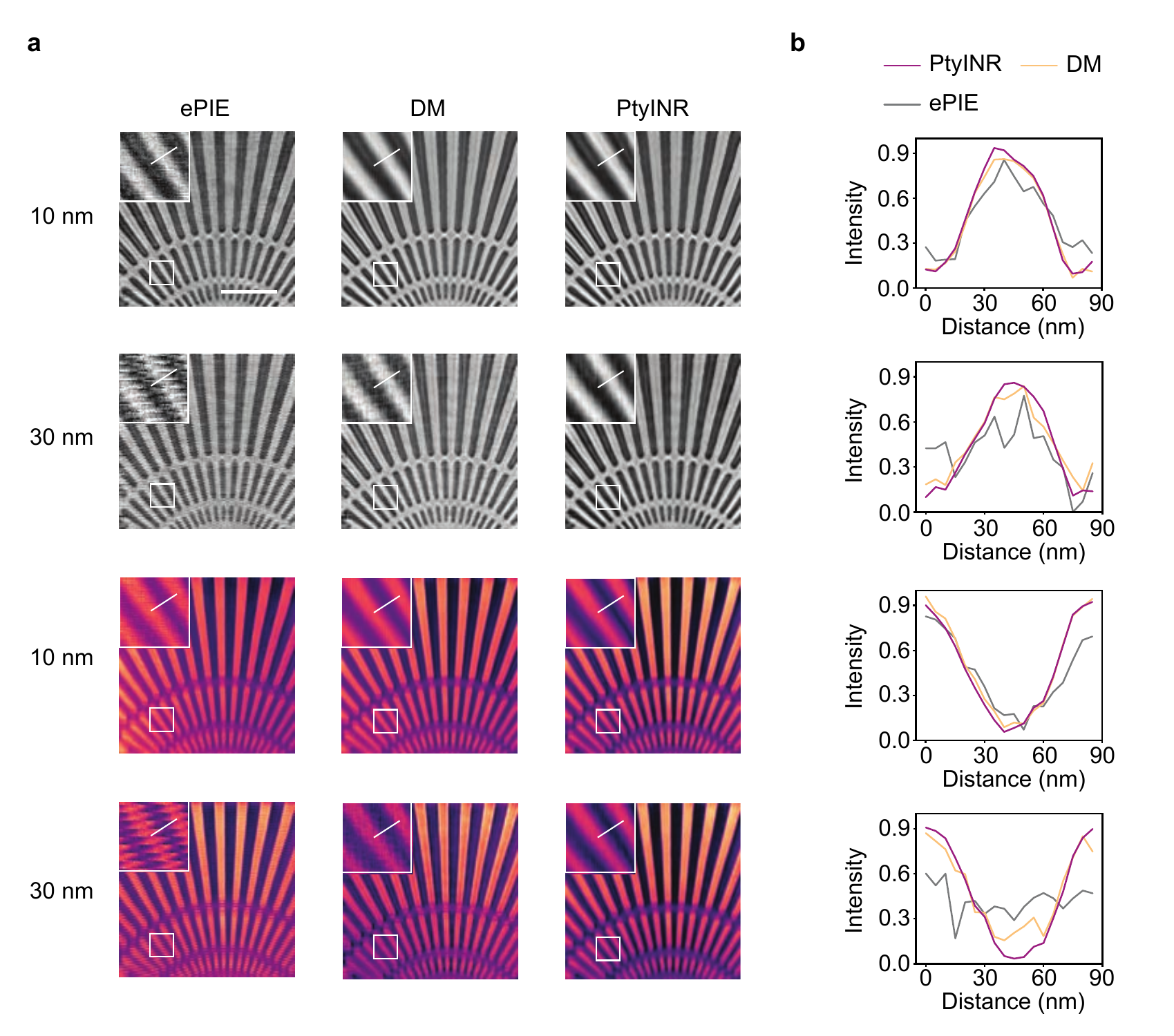}
    \caption{\textbf{Reconstructed object amplitude and phase corresponding to the Siemens-star test pattern dataset.}
    (\textbf{a}) Reconstructed object amplitude (top two rows) and phase (bottom two rows) using different methods (ePIE, DM, and PtyINR) at scanning step sizes of 10 nm and 30 nm. The results demonstrate the ability of each method to resolve fine spatial features in both amplitude and phase. Insets display magnified regions of interest with a 3× zoom factor. Scale bar: 600 nm.
    (\textbf{b}) Line profiles extracted from the regions marked in (a), comparing the resolved structural details across methods and step sizes. The profiles highlight differences in resolution and reconstruction fidelity.}
    \label{fig:supplementary_fig8}
\end{figure}
\clearpage

\section*{Supplementary Figure 10}
\begin{figure}[ht]
    \centering
    \includegraphics[width=\textwidth, height=\textheight, keepaspectratio]{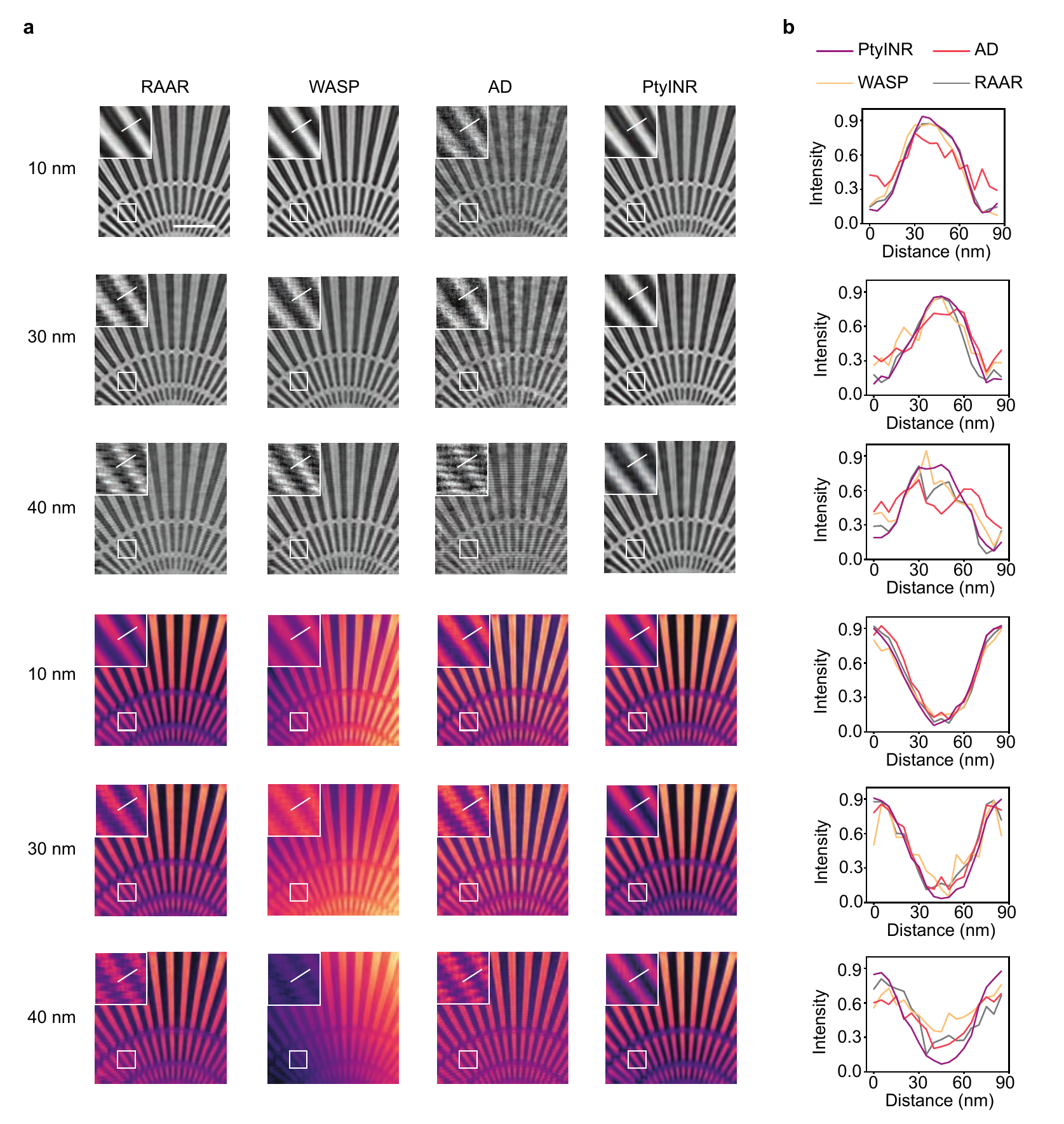}
    \caption{\textbf{Additional comparisons with reconstructed object amplitude and phase corresponding to the Siemens star test pattern dataset.}
    (\textbf{a}) Reconstructed object amplitude (top three rows) and phase (bottom three rows) using different methods (RAAR, WASP, AD, and PtyINR) at scanning step sizes of 10 nm, 30 nm, and 40 nm. The results illustrate the performance of each method in resolving fine structural details under varying spatial sampling conditions. Insets display magnified regions of interest with a 3× zoom factor. Scale bar: 600 nm. 
    (\textbf{b}) Line profiles extracted from the regions marked in (a), quantitatively comparing reconstruction fidelity and resolution across methods and step sizes in both amplitude and phase.}
    \label{fig:supplementary_fig9}
\end{figure}
\clearpage

\section*{Supplementary Figure 11}
\begin{figure}[ht]
    \centering
    \includegraphics[width=0.9\textwidth, height=0.9\textheight, keepaspectratio]{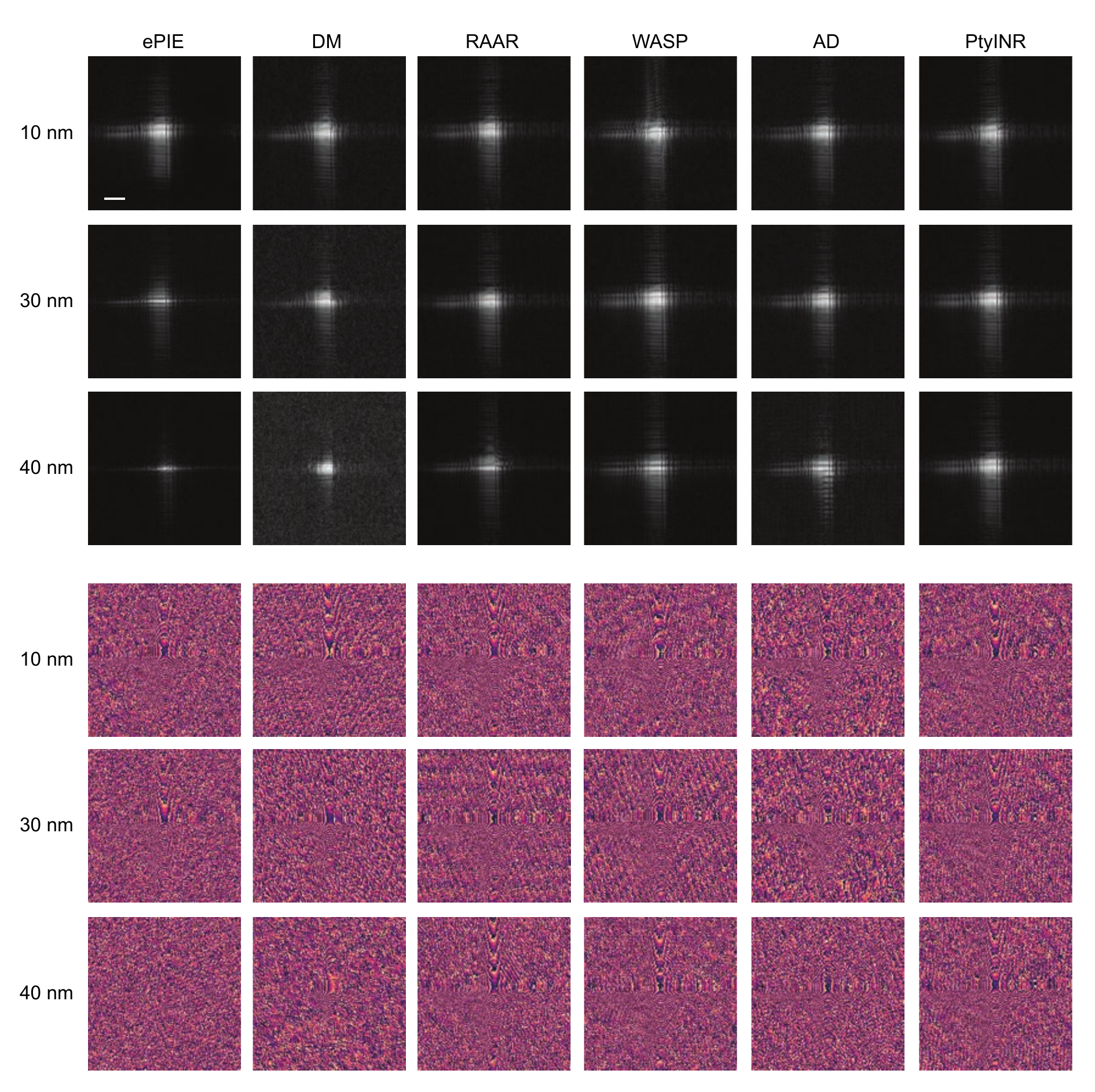}
    \caption{\textbf{Reconstructed probe functions corresponding to the Siemens-star test pattern dataset made of gold (Au).}
     Recovered probe amplitude and phase using different methods (ePIE, DM, RAAR, WASP, AD, and PtyINR) at scanning step sizes of 10 nm, 30 nm, and 40 nm. The comparison highlights the stability and consistency of probe recovery across methods and varying spatial sampling conditions. Scale bar: 150 nm.}
    \label{fig:supplementary_fig10}
\end{figure}
\clearpage

\section*{Supplementary Figure 12}
\begin{figure}[ht]
    \centering
    \includegraphics[width=0.9\textwidth, height=0.9\textheight, keepaspectratio]{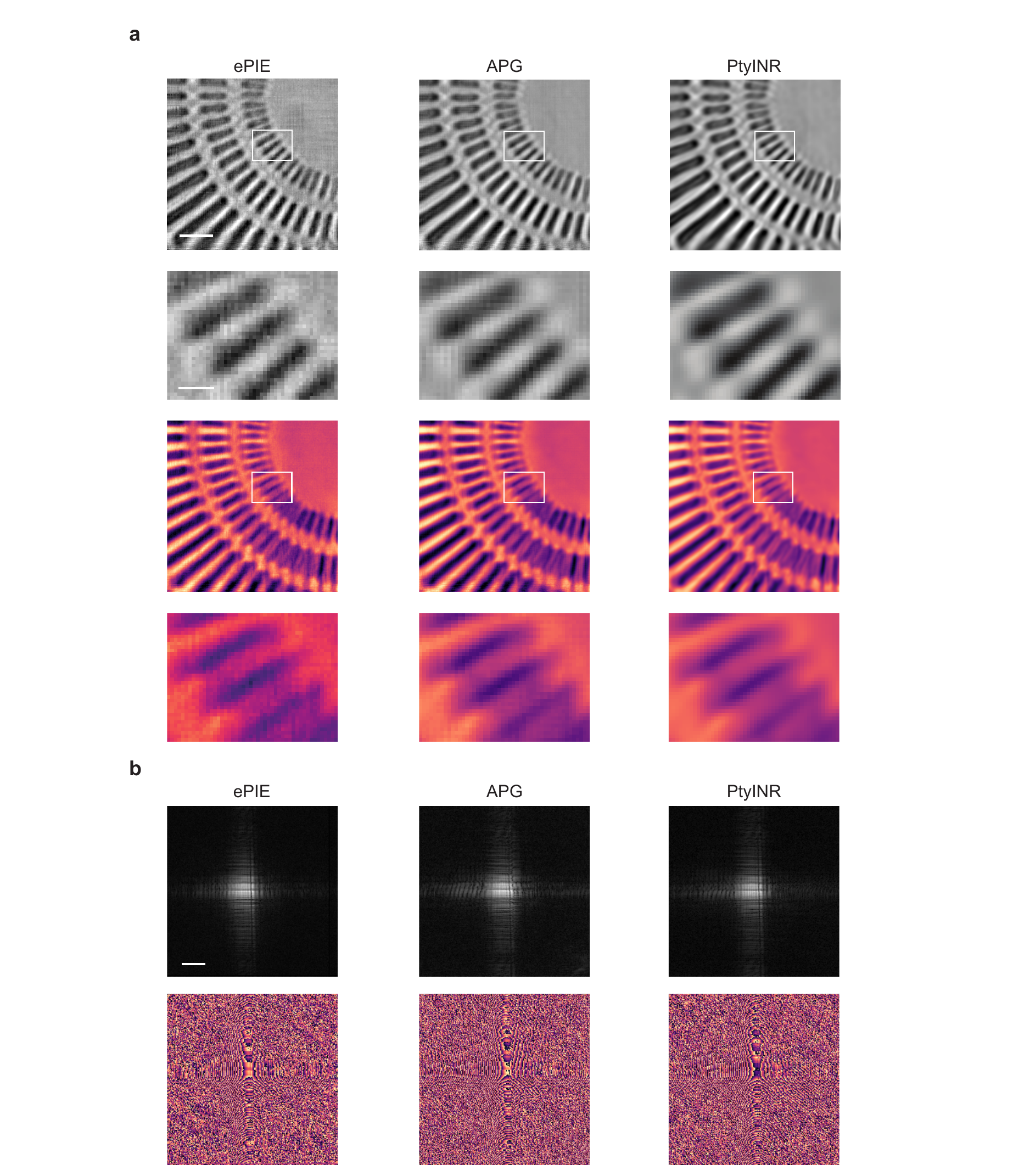}
    \caption{\textbf{Reconstruction results of the Siemens-star test pattern at an exposure time of 0.2 seconds.}
    (\textbf{a}) Reconstructed object amplitude (top two rows) and phase (bottom two rows) using different methods (ePIE, APG, and PtyINR). The first and third rows show the full-field reconstructions, while the second and fourth rows display zoomed-in views of selected regions from the amplitude and phase images, respectively, to better visualize fine structural details. Scale bar for the whole sample: 200 nm; Scale bar for extracted region: 50 nm;
    (\textbf{b}) Recovered probe amplitude (first row) and phase (second row) corresponding to each method, illustrating the stability of probe retrieval under high-exposure conditions. Scale bar: 150 nm.}
    \label{fig:supplementary_fig11}
\end{figure}
\clearpage

\section*{Supplementary Figure 13}
\begin{figure}[ht]
    \centering
    \includegraphics[width=0.95\textwidth, height=0.95\textheight, keepaspectratio]{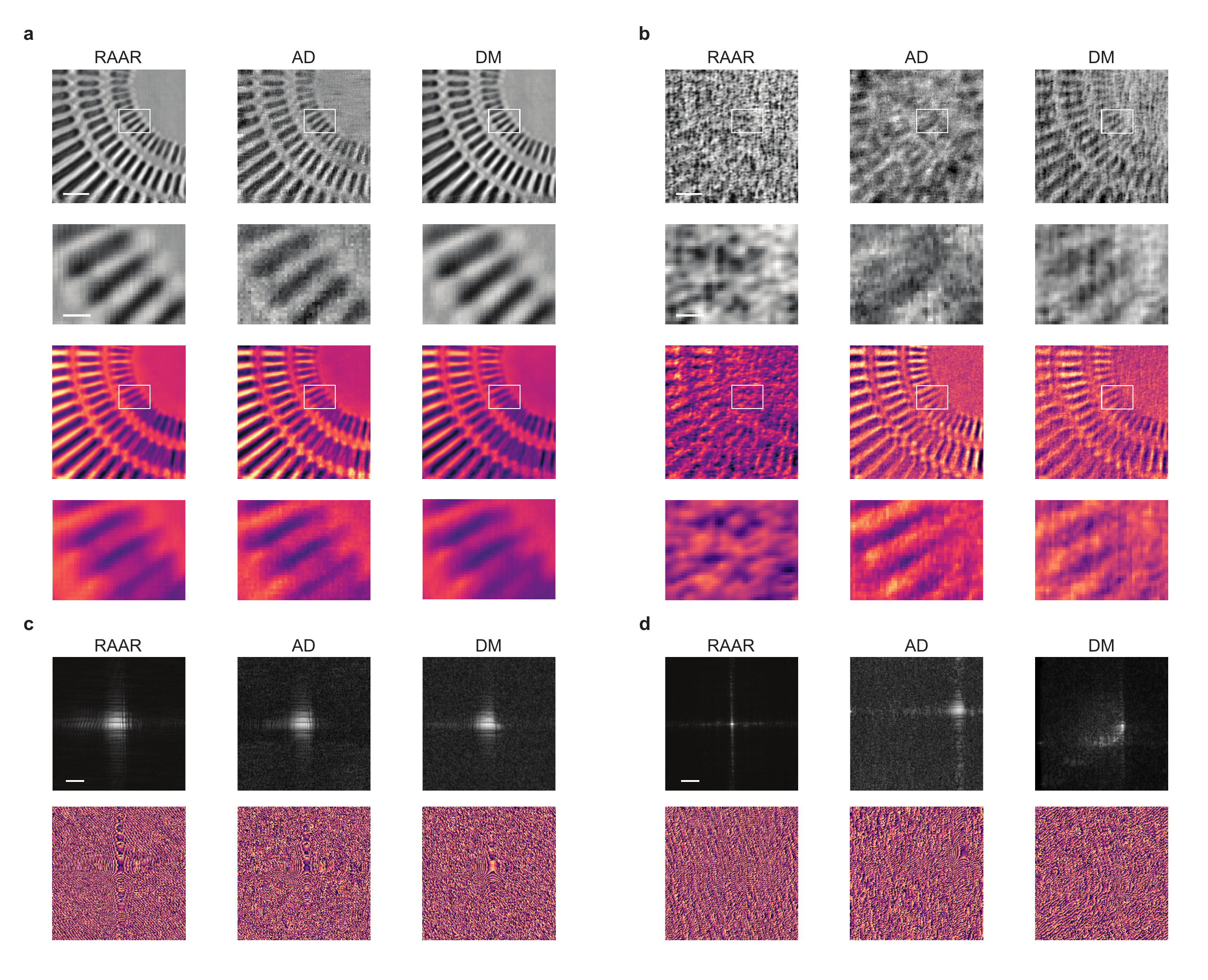}
    \caption{\textbf{Reconstruction results of the Siemens-star test pattern using RAAR and AD as well as DM methods at different exposure times.}
    (\textbf{a}) Reconstructed object at 0.2s exposure time. The top two rows show the object amplitude, and the bottom two rows show the phase. The second and fourth rows are zoomed-in views of the corresponding regions in the first and third rows, respectively. Scale bar for the whole sample: 200 nm; Scale bar for extracted region: 50 nm.
    (\textbf{b}) Reconstructed object at 0.003s exposure time, presented in the same layout as in (a), for comparison under low-exposure conditions. Scale bar for the whole sample: 300 nm; Scale bar for extracted region: 50 nm.
    (\textbf{c}) Recovered probe amplitude (top) and phase (bottom) corresponding to the 0.2s exposure time. Scale bar: 150 nm.
    (\textbf{d}) Recovered probe amplitude (top) and phase (bottom) corresponding to the 0.003s exposure time. The comparison illustrates the impact of reduced exposure on both object and probe reconstruction quality. Scale bar: 150 nm.}
    \label{fig:supplementary_fig12}
\end{figure}
\clearpage

\section*{Supplementary Figure 14}
\begin{figure}[ht]
    \centering
    \includegraphics[width=\textwidth, height=\textheight, keepaspectratio]{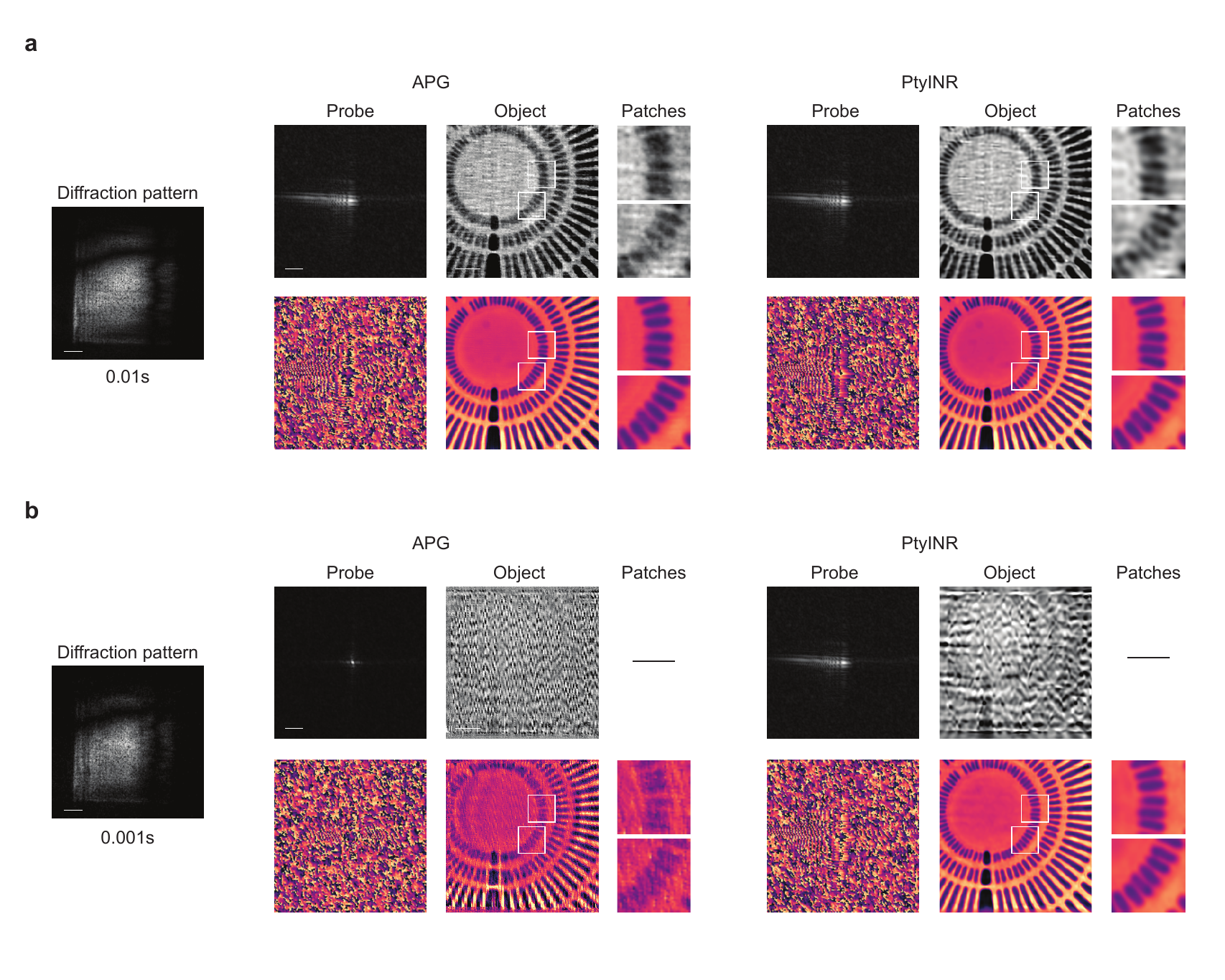}
    \caption{\textbf{Reconstruction results of the Siemens-star test pattern using APG and PtyINR methods at 0.01s and 0.001s exposure times.}
    (\textbf{a}) Leftmost column: measured diffraction pattern acquired with a 0.01 s exposure. The two panels to the right show reconstructions obtained with APG and PtyINR. In each reconstruction panel, the columns display (from left to right): reconstructed probe amplitude and phase; reconstructed object amplitude and phase; and selected patches from the reconstructed object corresponding to the regions marked by white rectangles.
    (\textbf{b}) The diffraction pattern, reconstructed probe, and reconstructed object at an exposure time of 0.001 s, presented in the same layout as in (a), demonstrating the robustness of PtyINR under low-dose conditions. Scale bar for the diffraction pattern: 1.5 mm; Scale bar for the reconstructed probe: 150 nm; Scale bar for the reconstructed object: 300 nm. The extracted patches correspond to a 2.8x magnification of the region outlined by the white rectangle.}
    \label{fig:supplementary_fig_add}
\end{figure}
\clearpage

\section*{Supplementary Figure 15}
\begin{figure}[ht]
    \centering
    \includegraphics[width=0.95\textwidth, height=0.95\textheight, keepaspectratio]{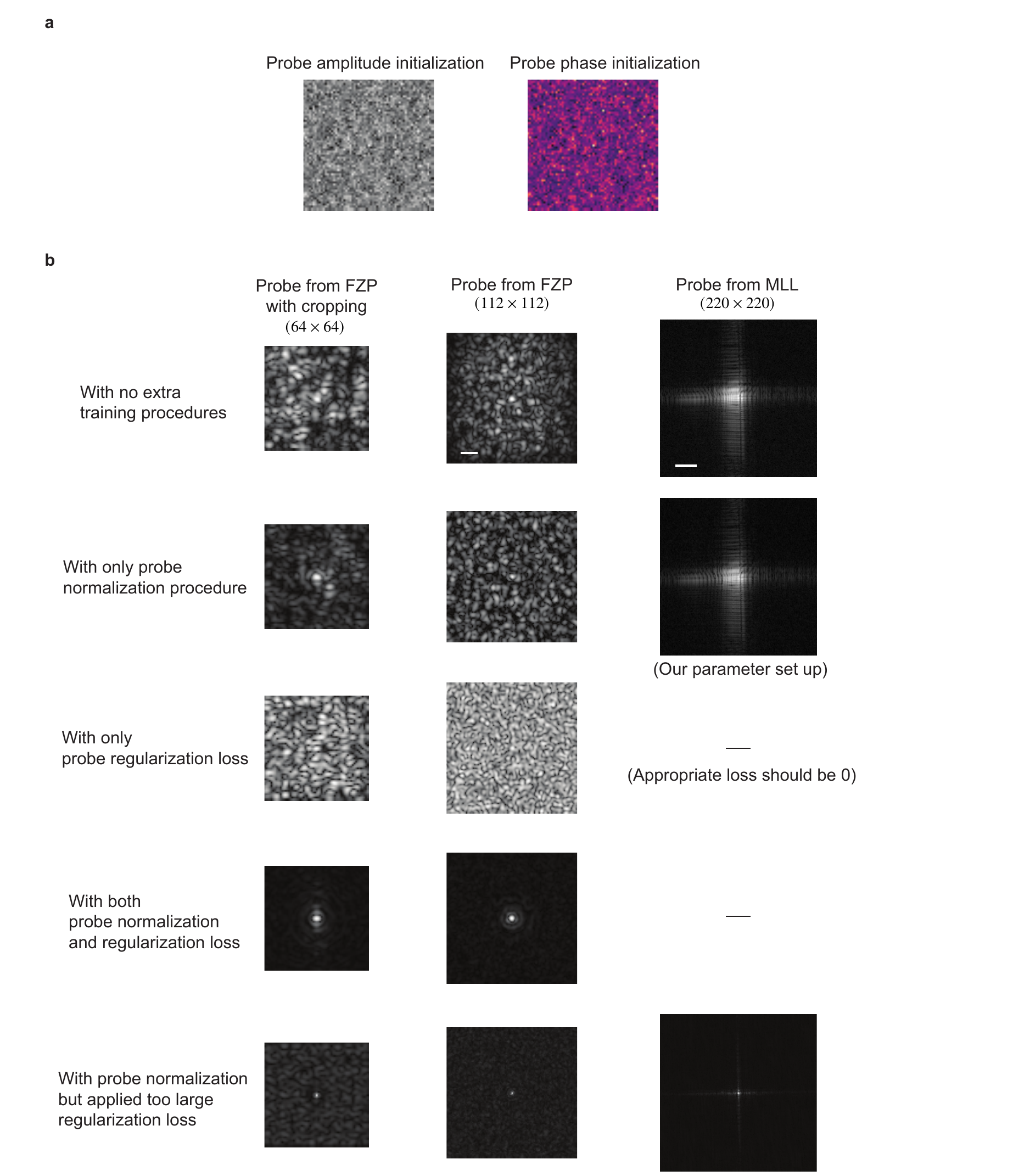}
    \caption{\textbf{Probe initialization and ablation study of probe recovery procedures for PtyINR.}
    (\textbf{a}) Probe initialization for PtyINR using uniform random values between 0 and 1 for both amplitude and phase.
    (\textbf{b}) Ablation study of probe recovery procedures under different configurations: without any additional operations, with only probe normalization ($P \leftarrow \frac{P}{\max(A_p)}$), with only regularization loss ($\lambda \bar{A_p}$), with both normalization and appropriate regularization, and with both but an overly strong regularization. These tests were conducted on three representative types of probes: a simple circular probe ($64\times64$) representing a basic FZP system, a more complex circular probe ($112\times112$) simulating an advanced FZP setup, and a rectangular probe ($220\times220$) modeled after a standard MLL focusing system. The results highlight the effectiveness and sensitivity of probe recovery to normalization and regularization choices. Scale bar for probe from FZP: 150 nm. Scale bar for probe from MLL: 150 nm.}
    \label{fig:supplementary_fig13}
\end{figure}
\clearpage

\section*{Supplementary Figure 16}
\begin{figure}[ht]
    \centering
    \includegraphics[width=0.75\textwidth, height=0.75\textheight, keepaspectratio]{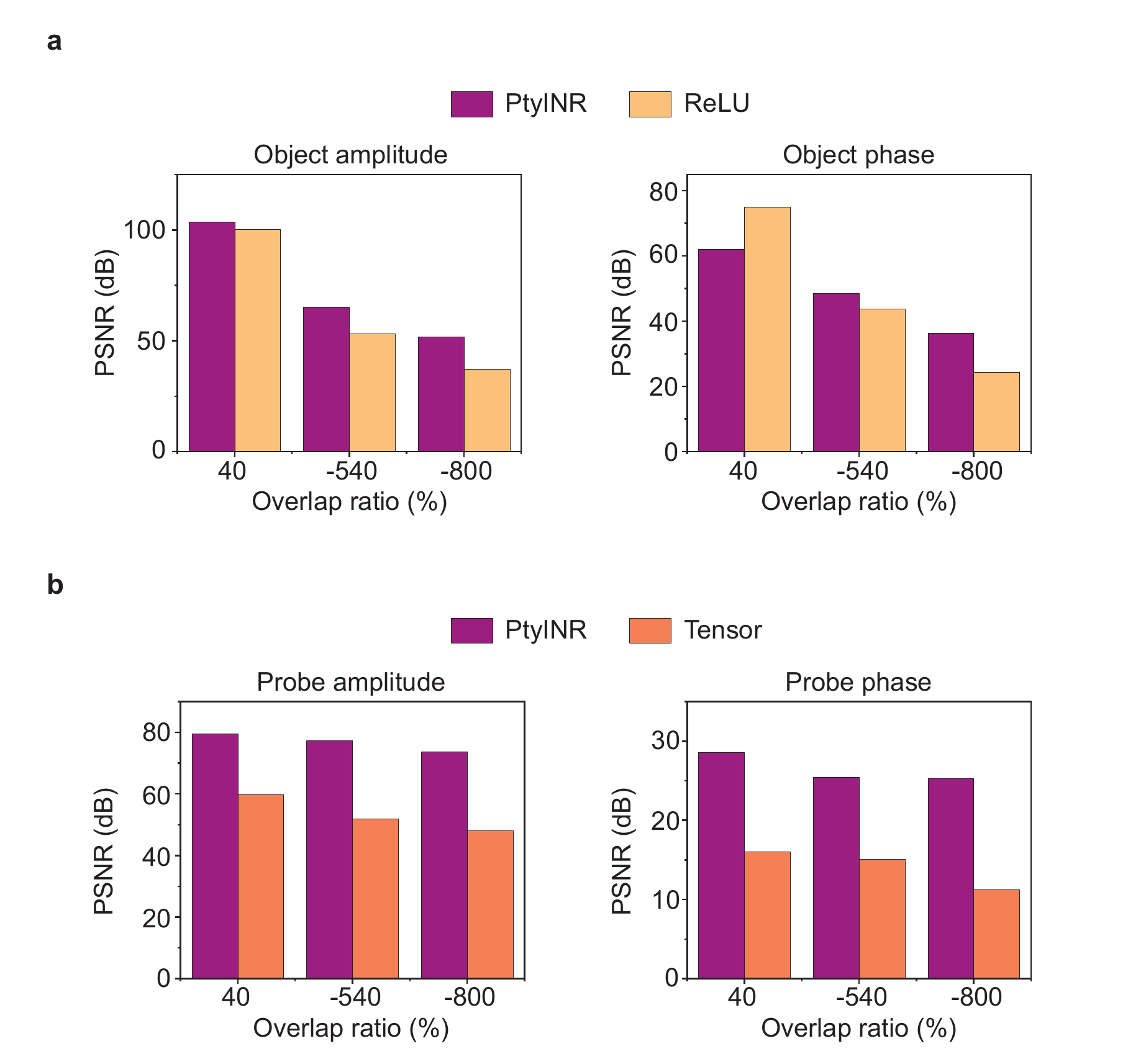}
    \caption{\textbf{The PSNR comparison of different modules in PtyINR for object and probe recovery under varying overlap ratios. }
    Histogram of PSNR values evaluated under different ptychographic overlap ratios (40\%, -540\%, and -800\%) to assess the effectiveness of neural network architectures in PtyINR.
    (\textbf{a}) Object recovery with a known probe: comparison between using the dedicated PtyINR object neural network and reusing the probe neural network (ReLU) architecture for object reconstruction.
    (\textbf{b}) Probe recovery with a known object: comparison between using the probe neural network and a direct tensor representation (non-network-based).
    The PSNR distributions clearly demonstrate that the tailored neural network architectures in PtyINR outperform mismatched or non-network alternatives, especially under low-overlap conditions.}
    \label{fig:supplementary_fig14}
\end{figure}
\clearpage

\section*{Supplementary Figure 17}
\begin{figure}[ht]
    \centering
    \includegraphics[width=0.9\textwidth, height=0.9\textheight, keepaspectratio]{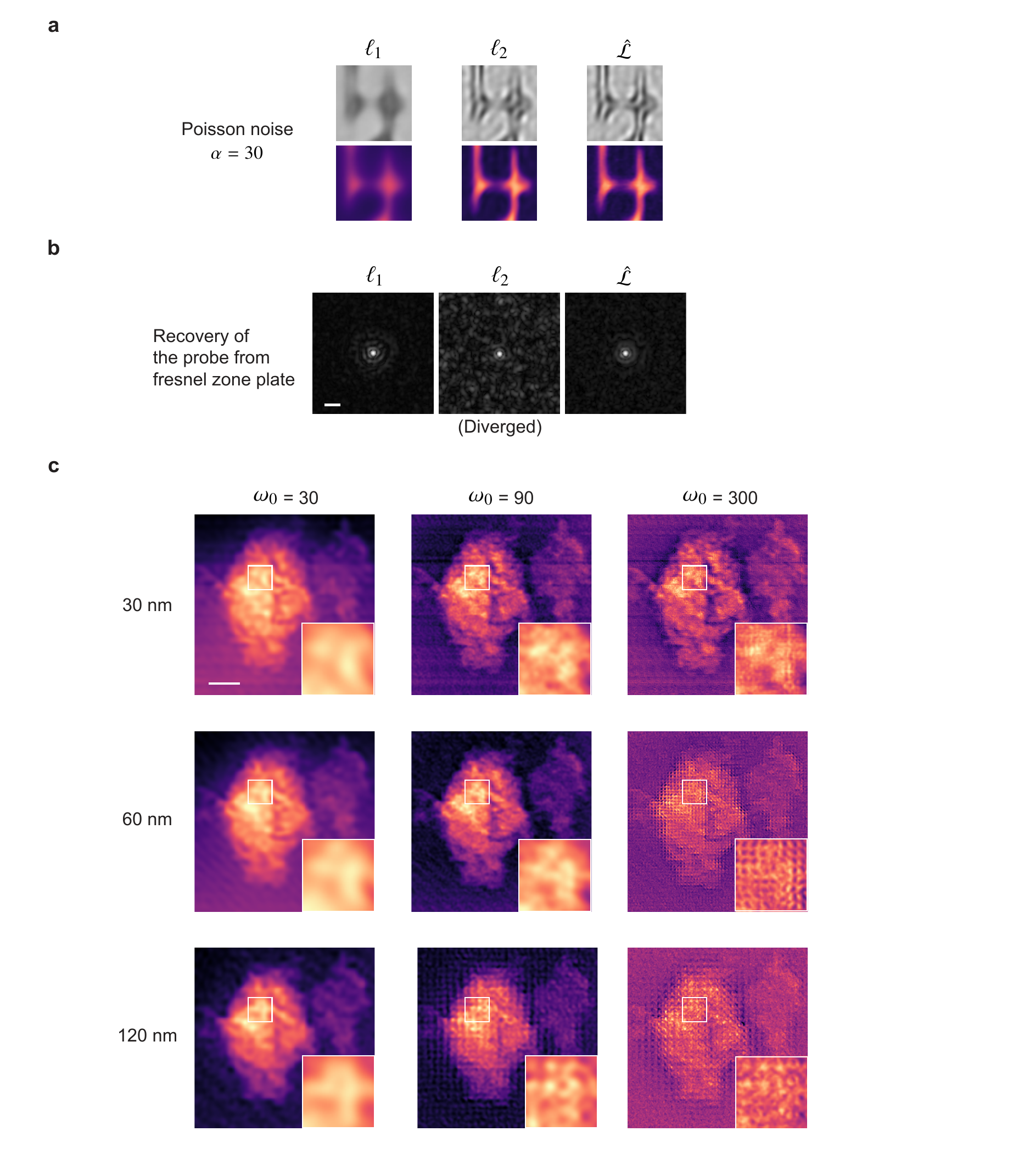}
    \caption{\textbf{Evaluation of loss functions and first omega parameter in PtyINR.}
    (\textbf{a}) Object reconstruction performance using different loss functions—$\ell_1$, $\ell_2$, and $\hat{\mathcal{L}}$ (as adopted in PtyINR)—under Poisson noise conditions. The comparison highlights the robustness of each loss function in recovering object amplitude and phase.
    (\textbf{b}) Recovery of complex FZP probe using the three loss functions. The recovered probe amplitudes and phases reveal the stability of the loss function choice on accurate probe modeling. Scale bar: 150 nm.
    (\textbf{c}) Influence of the first omega parameter in PtyINR on the final reconstruction quality under varying data acquisition conditions, including scanning step sizes of 30 nm, 60 nm, and 120 nm. Scale bar: 600 nm.}
    \label{fig:supplementary_fig15}
\end{figure}
\clearpage



\end{document}